\documentclass[runningheads]{llncs}

\usepackage{eccv}

\usepackage{eccvabbrv}

\usepackage{graphicx}
\usepackage{booktabs}

\usepackage{soulpos}

\usepackage[accsupp]{axessibility}  %

\usepackage{hyperref}

\usepackage{orcidlink}

\usepackage{multirow}
\usepackage{amssymb} %
\usepackage{pifont} %
\usepackage{arydshln} %

\begin{document}

\title{CarFormer: Self-Driving with Learned Object-Centric Representations} 

\titlerunning{CarFormer}

\author{Shadi Hamdan\orcidlink{0000-0002-8966-2347} \and
Fatma G\"uney\orcidlink{0000-0002-0358-983X}}

\authorrunning{S.~Hamdan and F.~G\"uney}

\institute{Department of Computer Engineering, Koç University\\
KUIS AI Center\\
\email{\{shamdan17,fguney\}@ku.edu.tr}}

\maketitle

\newcommand{\Perp}{\perp\!\!\! \perp}
\newcommand{\bK}{\mathbf{K}}
\newcommand{\bX}{\mathbf{X}}
\newcommand{\bY}{\mathbf{Y}}
\newcommand{\bk}{\mathbf{k}}
\newcommand{\bx}{\mathbf{x}}
\newcommand{\by}{\mathbf{y}}
\newcommand{\bhy}{\hat{\mathbf{y}}}
\newcommand{\bty}{\tilde{\mathbf{y}}}
\newcommand{\bG}{\mathbf{G}}
\newcommand{\bI}{\mathbf{I}}
\newcommand{\bg}{\mathbf{g}}
\newcommand{\bS}{\mathbf{S}}
\newcommand{\bs}{\mathbf{s}}
\newcommand{\bM}{\mathbf{M}}
\newcommand{\bw}{\mathbf{w}}
\newcommand{\eye}{\mathbf{I}}
\newcommand{\bU}{\mathbf{U}}
\newcommand{\bV}{\mathbf{V}}
\newcommand{\bW}{\mathbf{W}}
\newcommand{\bn}{\mathbf{n}}
\newcommand{\bv}{\mathbf{v}}
\newcommand{\bwv}{\mathbf{wv}}
\newcommand{\bq}{\mathbf{q}}
\newcommand{\bR}{\mathbf{R}}
\newcommand{\bi}{\mathbf{i}}
\newcommand{\bj}{\mathbf{j}}
\newcommand{\bp}{\mathbf{p}}
\newcommand{\bt}{\mathbf{t}}
\newcommand{\bJ}{\mathbf{J}}
\newcommand{\bu}{\mathbf{u}}
\newcommand{\bB}{\mathbf{B}}
\newcommand{\bD}{\mathbf{D}}
\newcommand{\bz}{\mathbf{z}}
\newcommand{\bP}{\mathbf{P}}
\newcommand{\bC}{\mathbf{C}}
\newcommand{\bA}{\mathbf{A}}
\newcommand{\bZ}{\mathbf{Z}}
\newcommand{\bff}{\mathbf{f}}
\newcommand{\bF}{\mathbf{F}}
\newcommand{\bo}{\mathbf{o}}
\newcommand{\bO}{\mathbf{O}}
\newcommand{\bc}{\mathbf{c}}
\newcommand{\bm}{\mathbf{m}}
\newcommand{\bT}{\mathbf{T}}
\newcommand{\bQ}{\mathbf{Q}}
\newcommand{\bL}{\mathbf{L}}
\newcommand{\bl}{\mathbf{l}}
\newcommand{\ba}{\mathbf{a}}
\newcommand{\bE}{\mathbf{E}}
\newcommand{\bH}{\mathbf{H}}
\newcommand{\bd}{\mathbf{d}}
\newcommand{\br}{\mathbf{r}}
\newcommand{\be}{\mathbf{e}}
\newcommand{\bb}{\mathbf{b}}
\newcommand{\bh}{\mathbf{h}}
\newcommand{\bhh}{\hat{\mathbf{h}}}
\newcommand{\btheta}{\boldsymbol{\theta}}
\newcommand{\bTheta}{\boldsymbol{\Theta}}
\newcommand{\bpi}{\boldsymbol{\pi}}
\newcommand{\bphi}{\boldsymbol{\phi}}
\newcommand{\bpsi}{\boldsymbol{\psi}}
\newcommand{\bPhi}{\boldsymbol{\Phi}}
\newcommand{\bmu}{\boldsymbol{\mu}}
\newcommand{\bsigma}{\boldsymbol{\sigma}}
\newcommand{\bSigma}{\boldsymbol{\Sigma}}
\newcommand{\bGamma}{\boldsymbol{\Gamma}}
\newcommand{\bbeta}{\boldsymbol{\beta}}
\newcommand{\bomega}{\boldsymbol{\omega}}
\newcommand{\blambda}{\boldsymbol{\lambda}}
\newcommand{\bLambda}{\boldsymbol{\Lambda}}
\newcommand{\bkappa}{\boldsymbol{\kappa}}
\newcommand{\btau}{\boldsymbol{\tau}}
\newcommand{\balpha}{\boldsymbol{\alpha}}
\newcommand{\nR}{\mathbb{R}}
\newcommand{\nN}{\mathbb{N}}
\newcommand{\nL}{\mathbb{L}}
\newcommand{\nE}{\mathbb{E}}
\newcommand{\cN}{\mathcal{N}}
\newcommand{\cM}{\mathcal{M}}
\newcommand{\cR}{\mathcal{R}}
\newcommand{\cB}{\mathcal{B}}
\newcommand{\cL}{\mathcal{L}}
\newcommand{\cH}{\mathcal{H}}
\newcommand{\cS}{\mathcal{S}}
\newcommand{\cT}{\mathcal{T}}
\newcommand{\cO}{\mathcal{O}}
\newcommand{\cC}{\mathcal{C}}
\newcommand{\cP}{\mathcal{P}}
\newcommand{\cE}{\mathcal{E}}
\newcommand{\cI}{\mathcal{I}}
\newcommand{\cF}{\mathcal{F}}
\newcommand{\cK}{\mathcal{K}}
\newcommand{\cW}{\mathcal{W}}
\newcommand{\cY}{\mathcal{Y}}
\newcommand{\cX}{\mathcal{X}}
\newcommand{\cZ}{\mathcal{Z}}
\def\bgamma{\boldsymbol\gamma}

\newcommand{\specialcell}[2][c]{%
  \begin{tabular}[#1]{@{}c@{}}#2\end{tabular}}

\newcommand{\figref}[1]{\Fig~\ref{#1}}
\newcommand{\secref}[1]{Section~\ref{#1}}
\newcommand{\algref}[1]{Algorithm~\ref{#1}}
\newcommand{\eqnref}[1]{Eq.~\eqref{#1}}
\newcommand{\tabref}[1]{Table~\ref{#1}}

\newcommand{\rulesep}{\unskip\ \vrule\ }

\newcommand{\KLD}[2]{D_{\mathrm{KL}} \Big(#1 \mid\mid #2 \Big)}

\renewcommand{\b}{\ensuremath{\mathbf}}

\def\mc{\mathcal}
\def\mb{\mathbf}

\newcommand{\T}{^{\raisemath{-1pt}{\mathsf{T}}}}

\makeatletter
\DeclareRobustCommand\onedot{\futurelet\@let@token\@onedot}
\def\@onedot{\ifx\@let@token.\else.\null\fi\xspace}
\def\eg{e.g\onedot} \def\Eg{E.g\onedot}
\def\ie{i.e\onedot} \def\Ie{I.e\onedot}
\def\cf{cf\onedot} \def\Cf{Cf\onedot}
\def\etc{etc\onedot} \def\vs{vs\onedot}
\def\wrt{wrt\onedot}
\def\dof{d.o.f\onedot}
\def\etal{et~al\onedot} \def\iid{i.i.d\onedot}
\def\Fig{Fig\onedot} \def\Eqn{Eqn\onedot} \def\Sec{Sec\onedot} \def\Alg{Alg\onedot}
\makeatother

\newcommand{\xdownarrow}[1]{%
  {\left\downarrow\vbox to #1{}\right.\kern-\nulldelimiterspace}
}

\newcommand{\xuparrow}[1]{%
  {\left\uparrow\vbox to #1{}\right.\kern-\nulldelimiterspace}
}

\newcommand*\rot{\rotatebox{90}}
\newcommand{\boldparagraph}[1]{\vspace{0.15cm}\noindent{\bf #1:} }
\newcommand{\boldquestion}[1]{\vspace{0.2cm}\noindent{\bf #1} }

\newcommand{\sh}[1]{ \noindent {\color{blue} {#1}} } 
\newcommand{\ftm}[1]{ \noindent {\color{cyan} {#1}}}

\newcommand{\cmark}{\checkmark}%
\newcommand{\xmark}{\ding{53}}%

\definecolor{First}{HTML}{BDE6CD}%
\definecolor{Second}{HTML}{E2EEBC}%
\definecolor{Third}{HTML}{FFF8C5}%

\newcommand{\fst}[1]{\cellcolor{First}#1}
\newcommand{\snd}[1]{\cellcolor{Second}#1}
\newcommand{\trd}[1]{\cellcolor{Third}#1}

\begin{abstract}
The choice of representation plays a key role in self-driving. Bird's eye view (BEV) representations have shown remarkable performance in recent years. %
In this paper, we propose to learn object-centric representations in BEV to distill a complex scene into more actionable information for self-driving. We first learn to place objects into slots with a slot attention model on BEV sequences. Based on these object-centric representations, we then train a transformer to learn to drive as well as reason about the future of other vehicles. We found that object-centric slot representations outperform both scene-level and object-level approaches that use the exact attributes of objects. %
Slot representations naturally incorporate information about objects from their spatial and temporal context such as position, heading, and speed without explicitly providing it.
Our model with slots achieves an increased completion rate of the provided routes and, consequently, a higher driving score, with a lower variance across multiple runs, affirming slots as a reliable alternative in object-centric approaches.
Additionally, we validate our model's performance as a world model through forecasting experiments, demonstrating its capability to predict future slot representations accurately. %
The code and the pre-trained models can be found at \href{https://kuis-ai.github.io/CarFormer/}{https://kuis-ai.github.io/CarFormer/}.
\keywords{Intermediate Representations \and Self-Driving \and Object-Centric Learning}
\end{abstract}
    
\section{Introduction}
\label{sec:intro}

The task of urban driving requires understanding the dynamics between objects in the scene. In self-driving, a scene is observed by the ego-vehicle via its sensors, typically multiple cameras. Then, for each observation step, an action is predicted and performed by the vehicle. In this paper, we propose an auto-regressive transformer-based model to reason about the dynamics of the scene while learning to drive. Starting with language processing, transformers are the de facto architecture today for sequential modeling from video generation\cite{Yan2021CoRR, Ren2022CVPR, Micheli2023ICLR} to robotics tasks. In language modeling, transformers typically operate on discretized inputs, or tokens, which are vastly different than high-dimensional continuous inputs such as camera images as in the case of self-driving.

There are various representation spaces used in self-driving ranging from pixels and coordinates to stixels and 3D points \cite{Janai2020Foundations}. In recent years, efforts converged to the bird's eye view (BEV) representation \cite{Chen2024Survey} which provides a top-down summary of the scene with scene elements that are relevant to driving such as lanes and vehicles. It enables the agent to concentrate on relevant semantics, free of distractions. %
However, despite its benefits, extracting accurate BEV maps from images has proven to be a challenging problem ~\cite{Philion2020ECCV, Li2022ECCV, Harley2022ICRA}. %
Therefore, we assume access to ground-truth BEV and start by discretizing it with a VQ-VAE to provide it as input to the transformer.
Although BEV provides a summary compared to six high-resolution camera images, it is still very high-dimensional. For example, most pixels belong to the road region and vehicles cover a relatively small portion of the BEV map despite being the primary cause of infractions. %
Our initial experiment with discretized BEV resulted in high infractions, motivating us to shift our focus to object-centric representations.

In object-centric approaches~\cite{Wu2023ICLR, Renz2022CORL}, the scene is represented in terms of objects for better modeling of scene dynamics. 
In simpler robotics environments~\cite{Wu2023ICLR}, this is achieved by placing objects into slots which then serve as tokens to model the interactions between objects. %
However, in the context of driving, learning to place objects into slots from driving sequences poses a significant challenge~\cite{Elsayed2022NeurIPS}.
Previous work has addressed this challenge by representing objects with vectors containing exact object attributes, including position, size, heading, and speed~\cite{Renz2022CORL}.
In our approach, we learn to extract relevant information about objects from their spatial and temporal context in BEV sequences. This allows our model to decide how to represent objects without explicitly specifying a set of attributes that may be incomplete, or subject to variations depending on the scene.

In this paper, we propose a learned object-centric approach to self-driving with slots. In a two-stage approach, we first learn to extract slots from BEV sequences with slot attention from videos~\cite{Kipf2022ICLR}. As a result of the first step, ideally, each object is placed into a slot that contains relevant information about the object. Each slot is naturally self-contained as it is responsible for reconstructing the object that it binds to.
Slot extraction is trained with a self-supervised objective and, therefore presents a more feasible and scalable alternative to exact object attributes. Moreover, it holds promise for generalization to diverse objects beyond vehicles. In the experiments, we show that slots can infer necessary information for driving such as the speed and the orientation of the vehicles. We analyze the factors that affect slot extraction and behavior learning with slots and find that increasing the number of slots and enlarging small objects improve performance significantly.

In the second step, we use a transformer to jointly learn the driving and the scene dynamics based on the extracted slot representations. 
We propose to replace the causal attention mechanism typically used in autoregressive transformers with block attention to allow interaction between all objects and the route for better modeling of dynamics in the scene.
For supervision, we apply loss on both continuous actions predicted using a GRU as in PlanT~\cite{Renz2022CORL} and on quantized actions predicted by the transformer autoregressively in multiple steps.
We perform extensive evaluations in terms of design choices including block attention and the choice of action head. In comparison to exact object attributes, slots result in a better completion rate of provided routes, leading to a higher driving score. Importantly, variation across multiple runs is also significantly lower with slots, which is a sign of robustness to variations in the scene during online evaluation.
In addition to improved driving performance, we qualitatively demonstrate that our model indeed learns dynamics with accurate predictions of slot representations in future steps. 
Our contributions are, in summary:
\begin{itemize}
    \item A learned, self-supervised, object-centric representation for self-driving based on slot attention, that contains the information necessary for driving such as speed and orientation of vehicles without explicitly providing them.
    \item CarFormer, an autoregressive transformer, that can both drive and act as a world model, predicting future states.
    \item State-of-the-art performance in the privileged setting of Longest6 benchmark, outperforming exact object-level attributes.
\end{itemize}

\section{Related Work}
\label{sec:rw}
We first provide an overview of representation spaces that are commonly used in self-driving with a special focus on BEV. We then briefly summarize the progress in self-supervised object-centric methods. Finally, we compare our approach to the recent sequence modeling approaches to control problems in robotics.

\boldparagraph{Representation Spaces in Self-Driving}
Starting with hand-designed affordances~\cite{Chen2015ICCV, Sauer2018CoRL}, various representation types have been proposed for self-driving. One common representation is semantic segmentation, initially in 2D~\cite{Muller2018CORL, Sax2019CORL, Mousavian2019ICRA, Zhou2019SR}, sometimes coupled with object detection for efficiency~\cite{Behl2020IROS}, and more recently in BEV~\cite{Chen2019CORL, Zhang2021ICCV, Hanselmann2022ECCV, Chen2022CVPR, Hu2023CVPR}. Another representation space is the coordinate space that is commonly used in trajectory prediction~\cite{Chang2019CVPR}. A notable work by Wang \etal uses a 2D detector to estimate the 3D properties of objects to render them in BEV~\cite{Wang2019IROS}. In addition to position, PlanT~\cite{Renz2022CORL} also includes other driving-related information such as heading and speed in its representation of each object through an attribute vector.
A representation based on object coordinates enables methods to concentrate on object-to-object relationships. However, it often lacks rich semantic information and may not adequately capture the varying spatiotemporal contexts of objects. Our goal is to address these limitations by integrating learned object-centric representations.

\boldparagraph{Self-Supervised Object-Centric Representations}
In this work, we build on the progress in object-centric learning where the goal is to decompose the scene into objects. A common way of achieving this goal is to integrate inductive biases about objects into the architecture, typically in an auto-encoding paradigm. Beginning with slot attention~\cite{Locatello2020NeurIPS}, these techniques reconstruct the input with a set of bottlenecks in the latent space called \emph{slots}. Each slot is expected to bind to an object region with similar visual cues. In this work, we adopt the methodology proposed by SAVi~\cite{Kipf2022ICLR}, which extends slot attention to video by incorporating temporal dynamics.

The progress in this domain started with synthetic images~\cite{Johnson2017CVPR, Karazija2021ARXIV}, and has since shifted towards in-the-wild images ~\cite{Everingham2010IJCV, Lin2014ECCV} and real-world videos~\cite{Ochs2013PAMI, Perazzi2016CVPR, Xu2018ECCV, Yang2019CVPRb, Li2013CVPR}. However, the existing methods struggle due to the complexity of unconstrained scenarios and resort to reconstructing different modalities such as flow~\cite{Kipf2022ICLR}, depth~\cite{Elsayed2022NeurIPS} or motion segmentation masks~\cite{Bao2022CVPR, Bao2023CVPR}. %
While recent studies~\cite{Seitzer2023ICLR, Aydemir2023NeurIPS} show promising results by performing reconstruction in the feature space of self-supervised models~\cite{Caron2021ICCV}, extracting slots from complex driving sequences remains a significant challenge. In this work, we assume the availability of a BEV representation of the scene through time as BEV resembles synthetic sequences where these methods perform reliably.

\boldparagraph{Transformers for Sequential Modeling}
Transformers are commonly applied to sequential prediction tasks in vision, such as video generation~\cite{Yan2021ARXIV, Ren2022CVPR, Micheli2023ICLR, Nash2022ARXIV}.
Previous work in robotics formulates Reinforcement Learning as a sequence modeling problem~\cite{Chen2021NeurIPS, Janner2021NeurIPS, Furuta2022ICLR, Zheng2022ICML}. %
However, these methods are typically evaluated on tasks where the state space is low-dimensional or can be straightforwardly encoded into a single vector.
For instance, in Decision Transformer~\cite{Chen2021NeurIPS}, the state is assumed to be adequately represented with a single token, which is encoded into a single vector with a CNN in the case of visual inputs. Trajectory Transformer~\cite{Janner2021NeurIPS} processes continuous states and actions by discretizing every dimension separately. %
While this works for low-dimensional action spaces, it is infeasible for the high-dimensional state space of self-driving.
One potential solution that we explore is to employ a VQ-VAE to quantize the input image into a lower dimensional discrete representation~\cite{Esser2021CVPR}. %

We do not constrain the state space to being discretized. Instead, we employ a hybrid modality for the inputs where some features are continuous while others are discrete. Although we utilize continuous inputs and non-causal \textit{block} attention, we maintain the autoregressive capability to generate future rollouts and reason on them, a crucial aspect of transformer-based RL approaches~\cite{Janner2021NeurIPS}. Non-causal \textit{block} attention is similar to the non-causal transformer decoder introduced in \cite{Liu2018ICLR}, but we do not limit the block to appear solely at the beginning of the sequence.

\section{Background on Slot Extraction}
\label{sec:bg}

\begin{figure*}[t]
  \includegraphics[width=\linewidth]{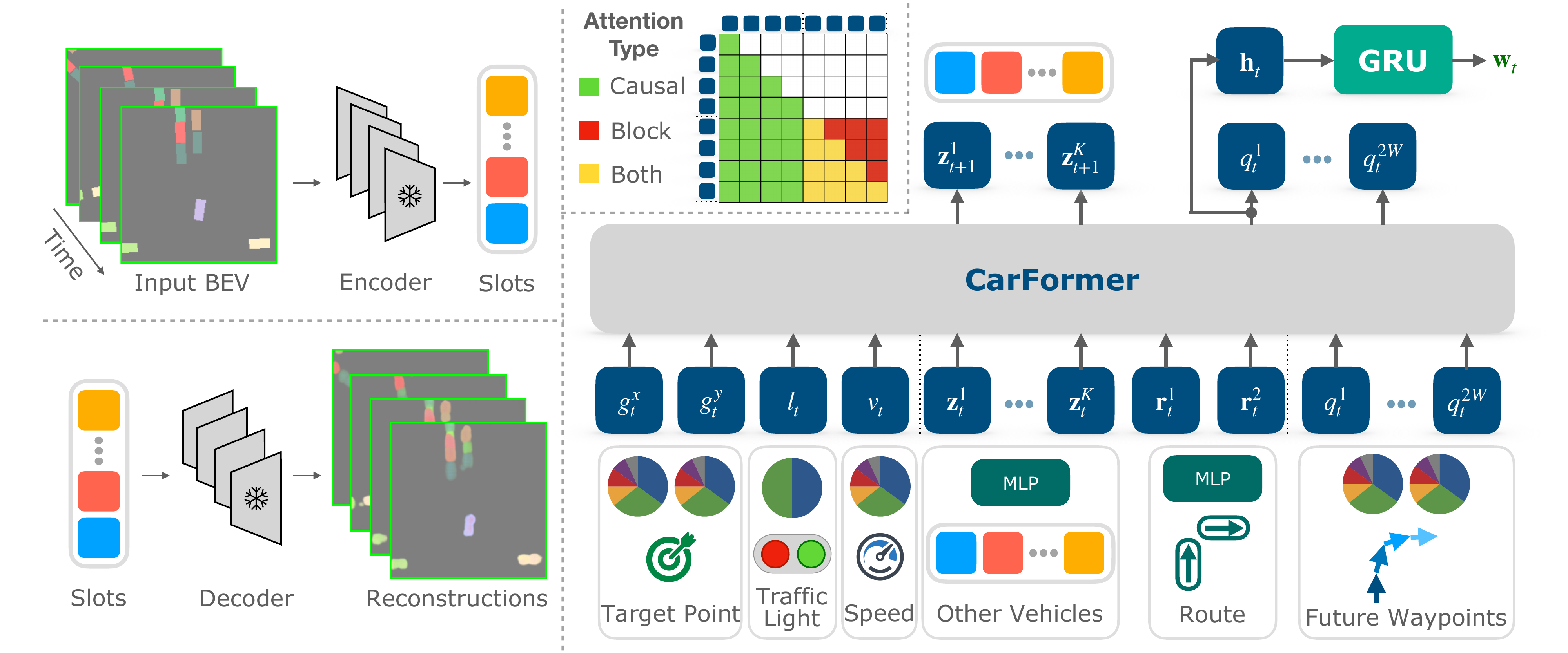}
  \caption{\textbf{Overview of CarFormer.} Given a trajectory $\tau_t$ consisting of discrete and continuous inputs, we first embed these tokens to the same hidden dimension $H$. For scalar inputs such as target point ($g_t^x$, $g_t^y$), traffic light flag $l_t$, speed $v_t$, and waypoints $q_t^i$, we discretize them, using k-means if not discrete, before we look them up from an embedding matrix. For continuous inputs like the $K$ slot features $\{\bz_t^i\}_{i=1}^K$ and the desired route $\br_t^1, \br_t^2$, we project them using an MLP. Conditioned on this context, CarFormer learns to jointly predict future slot features $\{\bz_{t+1}^i\}_{i=1}^K$ and the waypoints autoregressively using the backbone ($\bq_t$) as well as the GRU head ($\bw_t$). The K slot features $\{\bz_t^i\}_{i=1}^K$ are extracted from a pre-trained, frozen, SAVi model, shown on the left. The K slot features, along with the desired route, are considered a block, and block attention is applied in the attention layers as shown on the top.
  }
  \label{fig:overview}
\end{figure*}

Given the BEV representation of the scene in the last $T$ time steps, we first use a frozen object-centric model to extract the objects into slots. %
For extracting slots, we build on slot attention for videos, SAVi~\cite{Kipf2022ICLR}. Here, we provide a brief overview of SAVi for completeness and also, to introduce the notation with slots. Please see \cite{Kipf2022ICLR} for details of SAVi.

As a reconstruction-based method, SAVi follows an auto-encoding paradigm. Given a sequence of RGB frames $\bx_{t-T:t}$ with $T$ time steps for context, a CNN-based encoder is used to process each frame $\bx_i$ with positional encoding. The output of the encoder is flattened into a set of vectors $\bh_{t-T:t}$.
SAVi first initializes $K$ slot vectors $\tilde{\cZ}_{t-T} = \{\bz_{t-T}^{i}\}_{i=1}^K$. The initial set of slot vectors $\tilde{\cZ}_i$ is then updated with Slot Attention (SA)~\cite{Locatello2020NeurIPS} based on the visual features $\bh_i$ from the encoder for each time step $i$ within the interval, resulting in the updated set of slot vectors $\cZ_i$: 
\begin{equation}
    \cZ_i = f_{SA}\left(\tilde{\cZ_i}, \bh_i\right)
\end{equation}

To ensure temporally consistent slot representations, the subsequent slots are initialized based on the slot representation of the previous time step: $\cZ_{i+1} = f_{pred}\left(\cZ_{i}\right)$. 
In SAVi, slots are decoded into RGB predictions, and an alpha mask per slot for computing the reconstruction loss. We use the latent slot features $\cZ_t$ to describe time step $t$ in our model, \ie, without decoding.

\section{Methodology}
\label{sec:method}

We introduce CarFormer for learning to drive in the urban environment of CARLA~\cite{Dosovitskiy2017CORL}. Urban driving presents complexity due to the interactions between the ego-vehicle and other vehicles. Our goal is to learn driving behavior by capturing scene dynamics through slot representations. We formulate the behavior learning as a sequence modeling problem, as illustrated in \figref{fig:overview}. This sequence comprises tokens representing the goal, state, and action. We first define representations for each aspect before detailing the model architecture.

\subsection{Goal and State Representation} 
\label{sec:state}
We encode the route to follow and the state of the world in terms of tokens, which can be continuous or discrete, and feed them into our model. Specifically, we provide the model with the next target point in the route $(g_t^x, g_t^y)$, a flag signifying whether the ego vehicle is affected by a red traffic light $l_t \in \{0,1\}$, and the current speed of the ego vehicle $v_t \in \nR^+_0$. For each of these attributes, we apply k-means clustering and quantize them into $k_{attr}$ bins. %
For scene representation, we initially consider a scene-level representation by directly encoding the BEV map, and then we explore object-level representations.

\boldparagraph{Scene-Level Representation}
The input consists of a BEV representation of the scene at time $t$, denoted as $\bB_t \in {0, 1}^{192 \times 192 \times 8}$, centered around the ego-vehicle at time $t$ (\ie we use an ego-centric coordinate frame at each time-step). Each of the $8$ channels represents a binary map corresponding to a semantic class, such as road and vehicles~\cite{Zhang2021ICCV}.
Following common practice \cite{Yan2021ARXIV}, we use a VQ-VAE~\cite{Oord2017NeurIPS} to encode $\bB$ into a grid of discrete integers: $\bb_t \in \{1, \dots, C\}^{12 \times 12}$ where $C$ denotes the codebook size. We then flatten this grid to obtain a representation of the scene as a set of discrete tokens. Despite successful reconstructions, our experiments show that learning successful driving behavior on top of these discretized tokens is challenging (see \tabref{tab:sota}).

\boldparagraph{Object-Level Representations}
An alternative approach to representing the entire scene as a rasterized BEV is to represent individual objects within the scene. In PlanT~\cite{Renz2022CORL}, both vehicles and the desired route are represented as 6-dimensional vectors representing essential information such as size, position, orientation, and speed. Following PlanT, we initially train our model using the exact attributes of objects to represent the scene. This allows us to attribute improvements directly to the proposed object-centric representation with slots rather than our modifications to the transformer, such as block attention. %

In this paper, we propose an alternative object-level representation based on slots. %
Building on advancements in self-supervised object-centric representations, we explore slots as a more natural way of representing objects compared to exact object attributes. 
With slots, objects can be implicitly represented with relevant information from the spatio-temporal context of the object. %
Given the BEV representation of the scene in previous $T$ time steps, denoted as $\bB_{t-T:t}$, we employ SAVi~\cite{Kipf2022ICLR} to extract objects into $K$ slots, $\{\bz_t^i\}_{i=1}^K$ where each $\bz_t^i \in \nR^{1 \times d}$ is a $d$-dimensional slot vector corresponding to an object in the scene. As object-level representations do not include any information on the desired route, we also provide the model with the desired route represented by two vectors $\br_t^1, \br_t^2 \in \nR^6$ following PlanT \cite{Renz2022CORL}.

\subsection{Action Representation} 
Following common practice in self-driving~\cite{Renz2022CORL, Hanselmann2022ECCV, Chitta2023PAMI}, we predict waypoints that are then used to calculate the corresponding control consisting of throttle, brake, and steering angle. We predict waypoints in two ways: 

\boldparagraph{GRU} One way is to use a small GRU~\cite{Cho2014EMNLP} as in PlanT~\cite{Renz2022CORL}
The GRU is fed with the last latent vector from the backbone, concatenated with a flag representing the traffic light status.
With the GRU head, we sequentially predict $W$ future waypoints $\bw_t = \{(w_{x,i}, w_{y,i}) \}_{i=t+1}^{t+W}$ one by one in an autoregressive manner.

\boldparagraph{Quantization} Another way is to use our autoregressive transformer backbone and treat waypoint prediction as a next-word problem, a common strategy in language modeling approaches for RL~\cite{Chen2021NeurIPS, Janner2021NeurIPS, Shafiullah2022NeurIPS}. To implement this, we quantize the 2D waypoints by using k-means clustering on each dimension separately. As a result, $W$ waypoints are represented by a vector $\bq_t \in \{1, \dots, k_q\}^{2W \times 1}$, and then each dimension is predicted sequentially, conditioned on the previous predictions.

\subsection{CarFormer}
\label{sec:carformer}
Our goal is to learn self-driving in urban environments while jointly reasoning about scene dynamics as a sequence prediction task. %
At each time step, we represent the state of the world with %
a set of tokens as previously defined. %
We define a trajectory $\tau_t$ at time step $t$ as follows:
\begin{equation}
    \label{eq:traj}
    \tau_t = \{g_t^x, g_t^y, l_t, v_t, \bz_t^1, \dots, \bz_t^K, \br_t^1, \br_t^2, q_t^1, \dots, q_t^{2W}\}
\end{equation}
where $g_t^x, g_t^y$ denote the target point, 
$l_t$ the traffic light, $v_t$ the speed, $\{\bz_t^i\}_{i=1}^K$ object-level slot representations, and $\br_t^1, \br_t^2$ two attribute vectors representing the desired route. 
In PlanT style representation, slot vectors are replaced by the object attributes. 
In the case of scene-level representation, object-level slot representations and the desired route are replaced by the discrete tokens from the VQ-VAE, represented as $\{b_t^i\}_{i=1}^C$. 

\boldparagraph{Encoding} %
For each input in the trajectory, we handle discrete inputs, such as quantized target points and traffic light status, by performing a lookup from an embedding matrix. In the case of continuous attributes, like slot vectors, we project the vectors into the desired dimensionality using an MLP, as illustrated in \figref{fig:overview}. 
Specifically, for each discrete attribute, we initialize a $k_{attr}\times H$ embedding matrix, where $k_{attr}$ represents the number of possible values of the attribute after discretization, and $H$ denotes the hidden dimension of the backbone. Conversely, for continuous attributes, we utilize an MLP to project the vectors into $\mathbb{R}^H$.

\boldparagraph{Architecture} The backbone of the CarFormer is an autoregressive transformer decoder adapted from the architecture of GPT-2 \cite{Radford2019}. We modify the embedding layer to accommodate both continuous and discrete inputs simultaneously, enabling the incorporation of continuous representations such as slots while keeping other inputs like speed and traffic light discrete. %
Furthermore, we adjust the attention mechanism, which is causal in transformer decoders, to allow for cross-attention between certain blocks of the input. These blocks can be considered as a single unit, despite being composed of multiple tokens, such as the slot representations. To achieve this, we replace the triangular causal attention mask with a block triangular mask, enabling cross-attention within the block as shown on top in \figref{fig:overview}. We use this attention mechanism throughout our experiments and evaluate its impact in \tabref{tab:ablation}.

\subsection{Training}
\label{sec:training}
While our formulation can be extended to RL with rewards and multi-step predictions, currently, it corresponds to imitation learning with a single-step policy to predict action based on context. 
We train the model using imitation learning to learn to predict both the continuous waypoints $\bw_t$ and their quantized form, $\bq_t$ given the context as shown in \eqref{eq:traj}. We supervise both the GRU head, responsible for predicting continuous waypoints, and the language modeling head, responsible for predicting the quantized waypoints. To achieve this, we use the following loss functions:
\begin{eqnarray}
    \centering
    \label{eq:loss_action}
    \cL_{wp} &=& \cL_{\text{GRU}} + \cL_{\text{LM}}  \\
    \cL_{\text{GRU}} &=& \sum_{i=1}^W \lvert \bw_t^i-\hat{\bw}_t^i\rvert \nonumber \\
    \cL_{\text{LM}} &=& \sum_{i=1}^W \sum_{j=1}^2 \text{CE}(\bq_t^{i,j}, \hat{\bq}_t^{i,j}) \nonumber
\end{eqnarray}   

\boldparagraph{Auxiliary Forecasting}
Similar to PlanT~\cite{Renz2022CORL}, we additionally train the model to predict future scene representations jointly with action. In the case of the discretized scene representation using the VQ-VAE, we calculate the cross-entropy between the predicted logits and the ground truth future representation:
\begin{equation}
    \label{eq:loss_forecast_scene}
    \cL_{\text{scene}} = \sum_{i=1}^C ~\text{CE}(\bb_{t+f}^i, \hat{\bb}_{t+f}^i)
\end{equation}
where $f$ denotes the time horizon into the future for which we make predictions.

In the case of continuous scene representations, such as in object-level vectors or slot representations, we instead use the mean squared error between the representations: 
\begin{equation}
    \label{eq:loss_forecast_object}
    \cL_{\text{object}} = \sum_{i=1}^K ~\lVert \bz_{t+f}^i, \hat{\bz}_{t+f}^i \rVert
\end{equation}
The final loss is a weighted sum of the losses on action \eqref{eq:loss_action} and forecasting:
\begin{equation}
    \label{eq:total_loss}
    \cL_{wp} + \alpha~\cL_{\text{forecast}}
\end{equation}
where $\cL_{\text{forecast}}$ corresponds to $\cL_{\text{scene}}$ \eqref{eq:loss_forecast_scene} in case of scene-level and $\cL_{\text{object}}$ \eqref{eq:loss_forecast_object} in case of object-level. We experimentally set $\alpha$, the weight of forecasting, to $40$.

\section{Experiments}
\label{sec:exps}
\subsection{Experimental Setup}
\label{sec:exp_setup}
\boldparagraph{CARLA Setting}
We collect training data using the setup introduced in TransFuser~\cite{Chitta2023PAMI} on CARLA version 9.10.1., as also in PlanT~\cite{Renz2022CORL}. As PlanT~\cite{Renz2022CORL} shows that scaling the dataset size by collecting the data multiple times with different seeds leads to performance improvements, we adopt the $3\times$ setting, \ie collecting the data three times, as our default configuration. %
Our evaluation is conducted on the Longest6 benchmark, as proposed in TransFuser~\cite{Chitta2023PAMI}, comprising the six longest routes across six different towns. Further details on the expert driver and data collection routes can be found in TransFuser~\cite{Chitta2023PAMI}.

\boldparagraph{Data Filtering}
During data collection, we filter out expert runs that exhibit problematic behavior, typically occurring with specific seeds. We perform filtering by repeating runs where the expert achieves a driving score of less than $50\%$ with a different seed, ensuring consistency in dataset size. %
The filtered instances commonly involve vehicles spawned in orientations hindering movement, leading to expert timeouts, or scenarios failing to trigger, causing the expert to remain immobilized until timing out.

\boldparagraph{Evaluation Setting}
During online evaluations, unless stated otherwise, we employ the model with slots as input, $t+4$ as the target for future prediction, and the GRU head for waypoint prediction. Additionally, we implement creeping to prevent the agent from becoming stuck. %

\boldparagraph{Metrics}
When assessing driving performance, we report the metrics used in the CARLA leaderboard \cite{Dosovitskiy2017CORL}, namely \textit{Route Completion (RC)}, \textit{Infraction Score (IS)}, and \textit{Driving Score (DS)} as the combination of the two. For evaluating the predicted dynamics, we report the foreground version of the \textit{Adjusted Rand Index (ARI)} and mean \textit{Intersection over Union (mIoU)}. These metrics are computed by comparing predicted slot masks to target slot masks at a future time step.

\boldparagraph{Baselines}
Since we operate in the privileged setting, we benchmark against other privileged baselines. Specifically, we compare with AIM-BEV introduced in \cite{Hanselmann2022ECCV} and ROACH~\cite{Zhang2021ICCV} which use ground truth BEV as input. Additionally, we compare to PlanT~\cite{Renz2022CORL}, which incorporates ground truth object-level vehicles and routes as input. We obtain the results for AIM-BEV and ROACH directly from \cite{Renz2022CORL} and evaluate the publicly available PlanT-Medium checkpoint, trained on the 3$\times$ dataset on Longest6. %

\boldparagraph{Implementation Details}
Our dataset is divided into a training set ($94\%$ of the data), and validation and test sets ($3\%$ each). We train for 100 epochs on the training set and select the checkpoint with the best validation loss for online evaluation.
We set the hidden dimension of the transformer as $H=768$ and the number of layers to 6. %
In k-means, we use $k_g = 32$ for the target point, $k_v = 14$ for the speed, and $k_q = 48$ for the quantized waypoints.
For the target point and waypoints, we allocate half of the dimension for $x$ and the other half for $y$.
In SAVi, we set the number of context frames $T$ to 2 and the slot dimensionality $d$ to 128 and train it from scratch on a subset of our training split. Additional details can be found in Supplementary.

\begin{table*}[t]
\centering
\begin{tabular}{l c c c c} 
\toprule
\textbf{Model} & \textbf{Representation} & \textbf{DS$\uparrow$} & \textbf{IS$\uparrow$} & \textbf{RC$\uparrow$}\\ %
\toprule
CarFormer & \multirow{3}{*}{BEV} & 17.07{\small$\pm$3.78} & 0.30{\small$\pm$0.04} & 59.54{\small$\pm$4.34} \\ 
AIM-BEV$^*$~\cite{Hanselmann2022ECCV} & & 45.06{\small$\pm$1.68} & 0.55{\small$\pm$0.01} & 78.31{\small$\pm$1.12} \\
ROACH$^*$~\cite{Zhang2021ICCV} & & 55.27{\small$\pm$1.43} & 0.62{\small$\pm$0.02} & 88.16{\small$\pm$1.52} \\
\midrule
CarFormer & \multirow{2}{*}{Attributes} & 71.53{\small$\pm$3.52} & 0.78{\small$\pm$0.06} & 90.01{\small$\pm$1.60} \\
PlanT (Rep.)~\cite{Renz2022CORL} &  & 73.36{\small$\pm$2.97} & \textbf{0.84{\small$\pm$0.01}} & 87.03{\small$\pm$3.91} \\
\midrule
\multirow{1}{*}{CarFormer} & Slots & \textbf{\textbf{74.89{\small$\pm$1.44}}} & 0.79{\small$\pm$0.02} & \textbf{92.90{\small$\pm$1.28}} \\
\bottomrule
\end{tabular}
\caption{\textbf{Comparison on Longest6.} 
In the top part, we report the scene-level comparison where CarFormer significantly falls behind the other two approaches. In the middle, we report results using exact, object-level attributes, where CarFormer achieves a driving score within statistical significance of PlanT. With object-level slot representations shown in the bottom, CarFormer outperforms all scene-level approaches and even surpasses object-level approaches based on explicit object attributes. We report mean$\pm$std over 3 different runs. We report the results of PlanT reproduced (Rep.) using their official code. $^*$Results reported in PlanT~\cite{Renz2022CORL}.}
\label{tab:sota}
\end{table*}

\subsection{Quantitative Results}
\label{sec:quan_results}
\boldparagraph{Comparison} We present the results of online evaluations on the Longest6 benchmark 
in \tabref{tab:sota}. The table is divided into three based on the type of representation: scene-level representation at the top, followed by exact object-level attributes, and object-level slot representations at the bottom. In scene-level representations, CarFormer lags behind another imitation learning approach AIM-BEV~\cite{Hanselmann2022ECCV}, and an RL approach, ROACH~\cite{Zhang2021ICCV}. Despite accurately reconstructing input BEV with a VQ-VAE, the model cannot focus on objects, as evidenced by the significantly lower infraction score (IS). 

Compared to scene-level representation with VQ-VAE, we observe significant performance improvements with object-level representations. %
CarFormer with slots outperforms PlanT in RC despite a lower IS due to covering a longer distance, resulting in a higher mean DS with only half the variance compared to PlanT (\tabref{tab:sota}). This achievement is particularly noteworthy for two reasons: First, the slots model (bottom row) achieves this solely from BEV. While PlanT and CarFormer with attributes have access to exact agent locations, the slots model learns to accurately place agents in slots. Second, the significantly lower variance with slots demonstrates increased stability across runs, affirming slots as a more reliable alternative to attribute vectors. Note that the improved performance of our model cannot be attributed to architectural changes, as CarFormer with attributes performs worse than PlanT with higher variance.

\boldparagraph{Ablation Study} In \tabref{tab:ablation}, we perform an ablation study to evaluate our design choices when training and evaluating our model with slots as input. Specifically, we evaluate the effect of removing block attention, forecasting slots, and creeping when the vehicle is stuck during online evaluation. In the complete version of our model, we compare the results when using action predictions from either the GRU or the transformer itself with quantization (Q). %
We observed a significant improvement in results with the GRU and consequently adopted it for the remaining experiments.

\begin{table*}[t]
\setlength{\tabcolsep}{0.5em}
\centering
\begin{tabular}{l c c c c c c} 
\toprule
\textbf{Action} & \textbf{BA} & \textbf{Forecasting} & \textbf{Creeping} & \textbf{DS$\uparrow$} & \textbf{IS$\uparrow$} & \textbf{RC$\uparrow$}\\ %
\midrule

GRU & \cmark & \cmark & \cmark & \textbf{74.89{\small$\pm$1.44}} & 0.79{\small$\pm$0.02} & \textbf{92.90{\small$\pm$1.28}} \\
Q & \cmark & \cmark & \cmark & 66.87{\small$\pm$0.96} & 0.74{\small$\pm$0.01} & 87.12{\small$\pm$0.89}  \\ 
GRU & \xmark & \cmark & \cmark & 70.42{\small$\pm$2.68} & 0.78{\small$\pm$0.02} & 88.19{\small$\pm$2.00}  \\ 
GRU & \cmark & \xmark & \cmark & 57.25{\small$\pm$2.89} & 0.63{\small$\pm$0.03} & 89.54{\small$\pm$1.48} \\ 
GRU & \cmark & \cmark & \xmark & 69.16{\small$\pm$2.20} & \textbf{0.83{\small$\pm$0.01}} & 80.52{\small$\pm$1.88} \\ 

\bottomrule
\end{tabular}
\caption{\textbf{Ablation Study.} We first compare predicting continuous actions with the GRU to predicting quantized actions with the transformer~(Q). We also ablate the effect of removing Block Attention (BA) and slot forecasting and creeping. We present the results as the mean $\pm$ std over 3 different runs on Longest6.} 
\label{tab:ablation}
\end{table*}

The negative effect of removing creeping can be seen in a notably lower route completion score. Without creeping, the agent %
suffers from more frequent instances of getting stuck, leading to a decreased completion rate of assigned routes. %
This cautious behavior leads to a slightly higher IS but a lower overall DS. 
With BA, %
the agent can better model relationships between all objects in the scene due to the bi-directional attention within a block. As shown in the third row, removing BA results in increased infractions and a lower RC score. Adding forecasting significantly improves driving performance. The effect of forecasting is most visible in infraction scores, where removing it leads to a $0.17$ decrease in IS ($0.78 \to 0.63$).
This shows the importance of %
formulating the driving task jointly with forecasting, 
allowing the agent to anticipate the intentions of other vehicles and act accordingly. %

\boldparagraph{Slot Representations} When examining the performance of SAVi on BEV sequences, we identified two primary factors contributing to failure in slot extraction. Firstly, a low number of slots leads to missing vehicles as all slots become filled, particularly in crowded urban driving scenes. As demonstrated in \tabref{tab:slots}, the performance improvement from 7 to 30 slots clearly highlights the advantage of setting the number of slots high enough to capture majority of vehicles in the driving environment. However, adding more slots increases the computational load of training SAVi. To address this, we developed a lighter decoder to maintain efficiency while accommodating more slots. We provide an ablation of the light decoder in Supplementary.

\begin{table}[b]
\centering
\setlength{\tabcolsep}{0.5em}
\begin{tabular}{c c c c c} 
\toprule
\textbf{\#Slots} & \textbf{Enlarged} & \textbf{DS$\uparrow$} & \textbf{IS$\uparrow$} & \textbf{RC$\uparrow$}\\ 
\midrule
7 & \multirow{3}{*}{\xmark} & 48.17{\small$\pm$8.61} & 0.56{\small$\pm$0.08} & 86.70{\small$\pm$5.21} \\  %
14 &  & 49.34{\small$\pm$5.66} & 0.54{\small$\pm$0.07} & 92.33{\small$\pm$2.43} \\  %
30 &  & 71.48{\small$\pm$5.25} & 0.75{\small$\pm$0.06} & 93.00{\small$\pm$0.23} \\  %
\midrule
7 & \multirow{3}{*}{\cmark} & 62.93{\small$\pm$6.78} & 0.73{\small$\pm$0.07} & 80.20{\small$\pm$1.87} \\ %
14 & & 69.75{\small$\pm$8.03} & 0.78{\small$\pm$0.04} & 86.17{\small$\pm$2.79} \\ 
30 & & \textbf{74.89{\small$\pm$1.44}} & \textbf{0.79{\small$\pm$0.02}} & \textbf{92.90{\small$\pm$1.28}} \\  %
\bottomrule
\end{tabular}
\caption{\textbf{Effect of Slot Extraction on Driving.} We perform experiments to show the effect of enlarging small vehicles in order to overcome perception issues as well as varying the number of slots. We report mean$\pm$std over 3 different runs on Longest6.}
\label{tab:slots}
\end{table}

Increasing the number of slots leads to significant performance improvements, particularly in terms of infractions. Upon investigating the infractions with the lower number of slots, we discovered that most of them stemmed from small vehicles such as motorbikes or bicycles. Small vehicles, covering a relatively small area on the final BEV map, were often missed while decoding slots due to the small cost paid in terms of reconstruction loss. To test our hypothesis, we enlarged these small vehicles to increase their likelihood of being assigned to a slot. As shown at the bottom part of \tabref{tab:slots}, enlarging small vehicles notably enhances the performance, particularly with a small number of slots. The best performance is obtained with 30 slots and enlarging small vehicles.

\begin{table}[t]
\centering
\setlength{\tabcolsep}{0.75em}
\begin{tabular}{l c c c c c} 
\toprule
 & \multicolumn{2}{c}{\textbf{$T=t+1$}} & & \multicolumn{2}{c}{\textbf{$T=t+4$}} \\ 
\cline{2-3} \cline{5-6}
\textbf{Method} & \textbf{ARI$\uparrow$} & \textbf{mIoU$\uparrow$} & & \textbf{ARI$\uparrow$} & \textbf{mIoU$\uparrow$} \\ 
\midrule
Input-Copy &  0.641  &  0.561 & & 0.412 &  0.375 \\
CarFormer & 0.795  &  0.702 & & 0.540 &  0.454 \\
\addlinespace
\hdashline
\addlinespace
SAVi & 0.924  &  0.874 & & 0.924  &  0.874 \\ 
\bottomrule
\end{tabular}
\caption{\textbf{Forecasting Results.} We evaluate CarFormer's ability to predict future slots at time $T=T+1$ and $T=T+4$ that are decoded with the frozen SAVi decoder.} %
\label{tab:forecast}
\end{table}

\boldparagraph{Forecasting Future Slots} 
So far, we evaluated the performance of our model as a policy function to predict action. Additionally, our model can also predict future states of objects by learning to match the slot representations of SAVi in future time steps via the loss function defined in \eqref{eq:loss_forecast_object}. Therefore, CarFormer can be evaluated as a world model for predicting future states in addition to action. Similar to prior work on visual dynamics models~\cite{Wu2023ICLR}, we evaluate our model's performance in directly predicting future slot representations 1 or 4 timesteps ahead in \tabref{tab:forecast}. To be able to use object discovery metrics ARI and mIoU, we decode the predicted slot representations using the frozen decoder of SAVi. 

We report the performance of directly copying input as a sanity check (Input-Copy) and the performance of SAVi as an upper bound since SAVi has access to the current frame. 
Input-Copy works well for predicting the near future but degrades while predicting timestep $t+4$ due to objects moving away from their initial position. %
While there is a drop in our model's performance in predicting timestep $t+4$ as well, its predictions are more accurate in all metrics. Our model can learn the dynamics of objects in the scene and change their future position and orientation with respect to input.  %

\begin{figure*}[t]
    \centering
    \begin{minipage}{0.49\textwidth}
        \subfloat[]{\includegraphics[width=.99\linewidth]{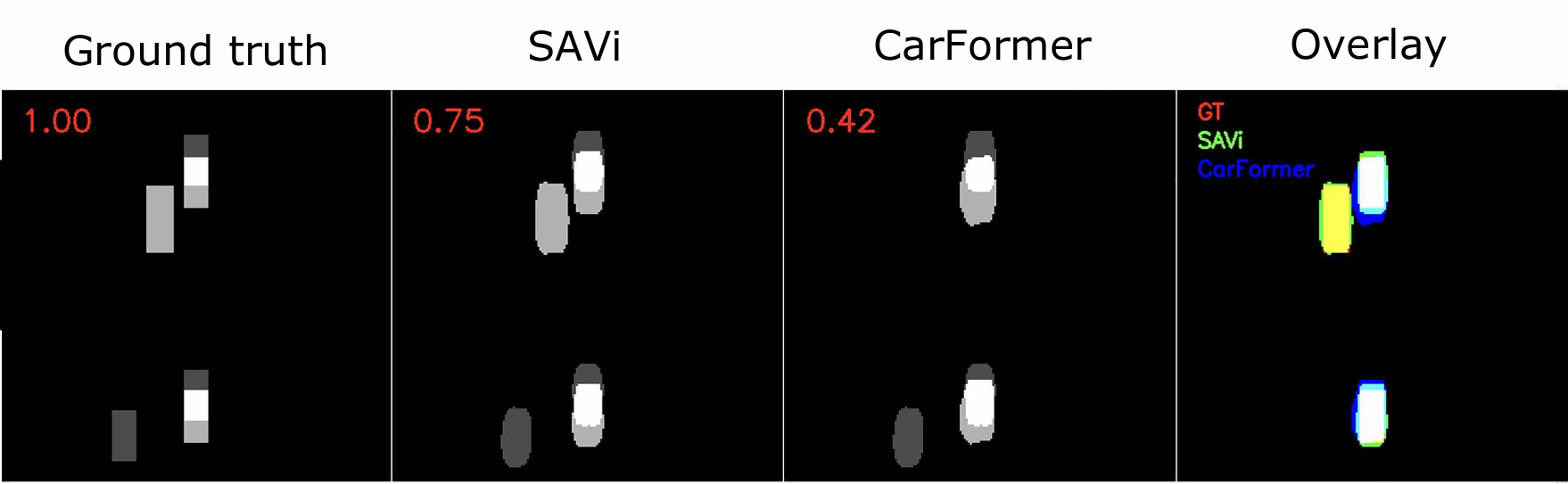} \label{fig:fa}} \\ %
        \subfloat[]{\includegraphics[width=.99\linewidth]{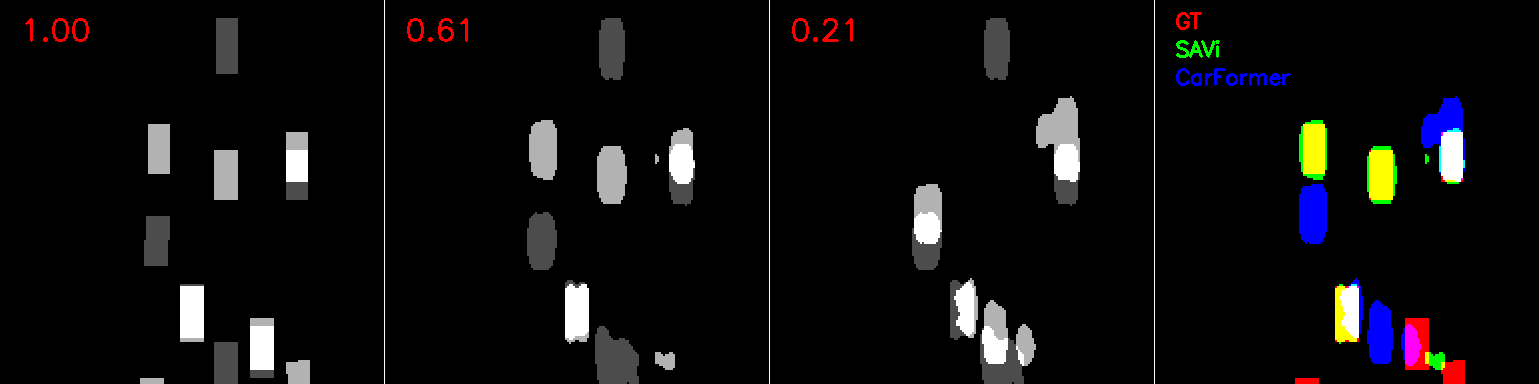} \label{fig:fb}} \\ %
        \subfloat[]{\includegraphics[width=.99\linewidth]{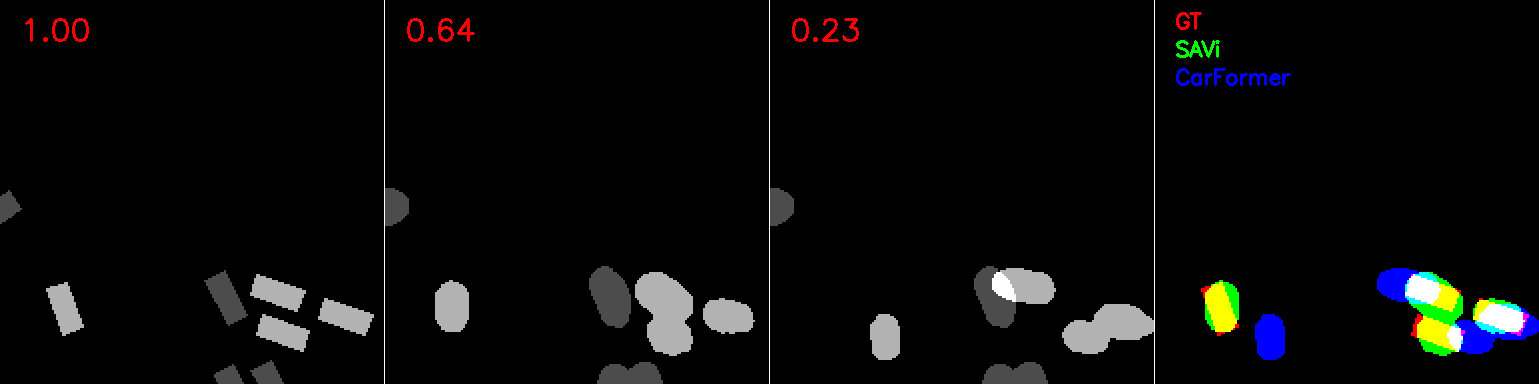} \label{fig:fc}}
    \end{minipage}%
    \hfill
    \begin{minipage}{0.49\textwidth}
        \subfloat[]{\includegraphics[width=.99\linewidth]{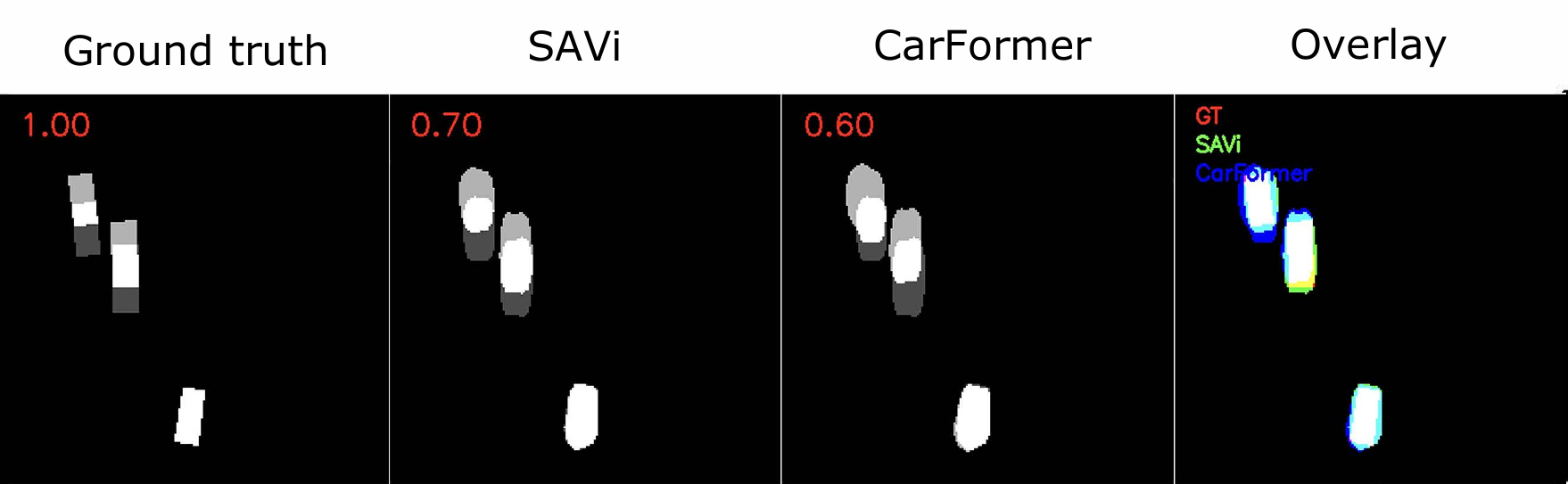} \label{fig:fd}} \\ %
        \subfloat[]{\includegraphics[width=.99\linewidth]{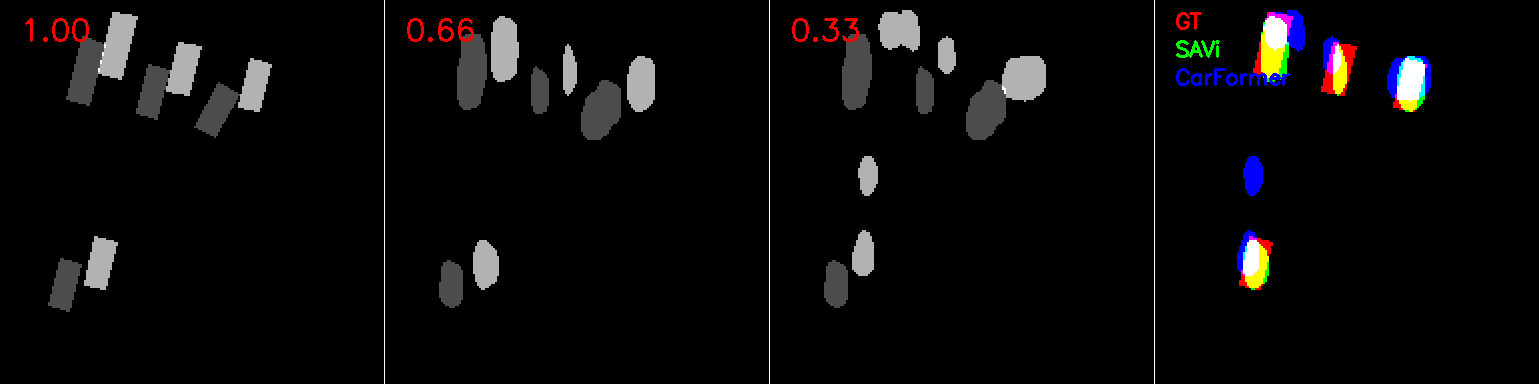} \label{fig:fe}} \\ %
        \subfloat[]{\includegraphics[width=.99\linewidth]{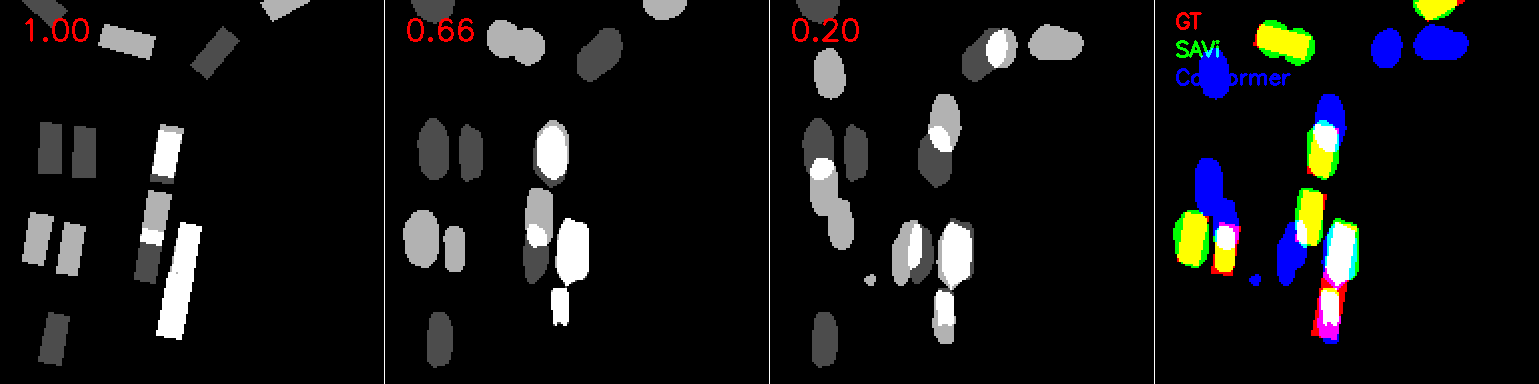} \label{fig:ff}} 
    \end{minipage}%
    \caption{\textbf{Visualization of Slot Forecasting Results.} Each sub-figure shows an example of input~(dark grey)-output~(light grey) objects in the first column, SAVi reconstructions in the second column, and our model's predictions in the third column. The top left corner of each column shows the mIoU compared to the ground truth. For comparison, we overlay the three in the last column where the red channel~(\textcolor{red}{R}) is the ground-truth location, the green channel~(\textcolor{green}{G}) is SAVi reconstruction, and the blue channel~(\textcolor{blue}{B}) is our prediction. In the case of perfect alignment between the three, we see the vehicles in white, and different errors for our model can be seen in a unique color %
    such as yellow~(\textcolor{yellow}{R+G}) indicating misses and blue indicating false positives~(\textcolor{blue}{B}).}
    \label{fig:forecast}
\end{figure*}

\subsection{Qualitative Results}
\label{sec:qual_results}
In \figref{fig:forecast}, we visualize predictions of our model (third column) in comparison to ground truth (first column) and SAVi reconstructions (second column) which we use to supervise our model for predicting future slot representations. In the first three columns, we display the initial locations of the vehicles in dark grey and the final locations that we aim to predict in light grey. In the last column, we overlay the final positions in all three columns, each represented in a different channel: ground truth in red, SAVi in green, and CarFormer in blue. %
This allows us to clearly identify various types of errors for our model in a specific color such as false negatives in blue and false positives in yellow. 

\boldparagraph{Proper Dynamics} As illustrated in examples \subref{fig:fa}, \subref{fig:fc}, \subref{fig:fd}, and \subref{fig:fe}, our model effectively learns visual dynamics between objects. In \subref{fig:fa}, our model recognizes that the vehicle in the bottom left is moving backward (downward), and accurately predicts its absence in the future. In the same example, our model correctly infers that vehicles on the right are heading forward at a slower speed than the ego vehicle, and thus accurately predicts their future locations. %
In \subref{fig:fd}, our model accurately identifies that all vehicles are heading forward, with those on the top-left moving slightly faster than the ego vehicle, thus they end up slightly further up. These examples demonstrate that our model can deduce vehicle speeds from slot representations, without explicit information on heading direction or speed. 
Its capability extends beyond simply associating left/right lanes with heading direction, as can be seen from \subref{fig:fa} and \subref{fig:fd}.
In case of a lane change (\subref{fig:fe}) or a turn (\subref{fig:fc}), our model accurately predicts the change in its viewpoint. 

\boldparagraph{Problems in Forecasting} We encounter three types of issues while forecasting slots. The first one occurs due to the perception errors by SAVi during slot extraction. Since we rely on slot representations extracted by SAVi both as input and also for supervision while forecasting slots, our model's performance is constrained by SAVi's performance. This can be observed from errors caused by blurry predictions of SAVi in turning scenarios in \subref{fig:fe} and \subref{fig:fc} or from failing to locate cars in crowded scenes in \subref{fig:fb}. The second type of problem involves False Positives, resulting from hallucinated vehicles, as indicated by the blue color in the last column of \subref{fig:fe}, \subref{fig:fc}, and \subref{fig:ff}. Lastly, our model struggles to predict complex dynamics when multiple vehicles are moving at different speeds as shown in \subref{fig:fc}.

\section{Conclusion and Future Work}
In this paper, we introduced CarFormer as the first approach to self-driving with object-level slot representations. We demonstrated that reasoning with slots not only improves the driving score but also provides robustness across variations in multiple online evaluations. 
We trained and validated the performance of CarFormer both as a policy to predict action and as a visual dynamics model for predicting future states of objects. %
Unlike PlanT, which utilizes a transformer encoder to process a single time step, we employed an autoregressive transformer decoder in CarFormer. This design has the potential to be extended to multi-step reasoning with reward/return tokens, as demonstrated in robotics tasks~\cite{Janner2021NeurIPS, Chen2021NeurIPS}. In comparison to robotics tasks, self-driving has a more complex state representation due to intricate dynamics between objects in addition to well-known challenges in appearance, especially when extracting information from cameras. 

We currently assume access to the ground truth BEV maps in our model. Despite significant progress in learning BEV representations in recent years, BEV perception still lacks the accuracy needed to extract slots from it in urban driving scenes. Instead of a two-stage approach by first estimating BEV and then extracting slots from it, a more direct approach of extracting slots in BEV might be preferable, both in terms of efficiency and avoiding cascading errors. %
With the advancements in slot extraction from real-world videos, any object that can be placed into a slot can be part of the reasoning in our model.

\boldparagraph{Acknowledgements} We thank Barış Akgün, Deniz Yuret, and members of AVG at Koç University for discussions and proof-reading, Hongyang Li and Li Chen for helpful suggestions during the rebuttal, and Katrin Renz for sharing details of PlanT. This project is co-funded by the KUIS AI Center and the European Union (ERC, ENSURE, 101116486). Views and opinions expressed are however those of the author(s) only and do not necessarily reflect those of the European Union or the European Research Council. Neither the European Union nor the granting authority can be held responsible for them.

\clearpage
\setcounter{page}{1}
\setcounter{section}{0}
\maketitlesupplementary

\begin{abstract}
In this Supplementary, we first provide the \textbf{details} about the architecture (\secref{sec:arch_details}) and training (\secref{sec:train_details})
for reproducibility. We then provide \textbf{additional quantitative results} in \secref{sec:add_quan} on ablating our decision to use a lighter architecture in slot extraction and on the effect of enlarging slots and block attention in terms of forecasting performance. We also report the \textbf{inference time and the number of parameters} of our model at the end of \secref{sec:add_quan}. We \textbf{visually compare} slot reconstructions between different versions of SAVi as well as forecasting results by our model to ground truth and SAVi in \secref{sec:add_qual}. Finally, we provide a road map for \textbf{future work} to autoregressively generate rollouts with block attention in \secref{sec:rollout_gen}.
Please also see the \textbf{video} in the zip for a visualization of our model's driving performance.
\end{abstract}

\section{Architectural Details}
\label{sec:arch_details}
\boldparagraph{Slot Extraction}
To extract slots, we build on SAVi~\cite{Locatello2020NeurIPS} based on its implementation in \cite{Wu2023ICLR}. Additionally, we process the input BEV representations by assigning a different color to each vehicle, which we found to help in slot extraction. To color the vehicles, we sample a random color per vehicle from a set of 14 colors and fix this color by the vehicle ID. To use enlargement, we enlarge any vehicles that are less than $4.9m \times 2.12m$ to $4.9m \times 2.12m$. In the absence of privileged information, we detect vehicles in BEV space by finding connected components in the BEV occupancy grid. Then, to enlarge vehicles, we continuously dilate the connected component until we surpass an area threshold. We experimentally set this area threshold to $8m^2$, corresponding to 200 pixels in our $5$ pixels per meter setting. To randomly color objects, we similarly assign a random color per connected component. To maintain color per vehicle across time steps, we compare the current BEV to the previous time step and reuse the color of the previous connected component if the previous connected component covers over $50\%$ of the area of the current connected component.

We provide the hyper-parameters for different versions of the model in \tabref{tab:saviparams}. Due to computational requirements increasing with the number of slots, we proposed a lighter version of the base SAVi model we start with. In the light version, we reduced the number of parameters in the decoder and the encoder MLP. 
Using the lighter version, we train SAVi with up to 30 slots in our experiments.
We highlight the differences between the versions in bold in \tabref{tab:saviparams}.

\begin{table}[ht]
    \setlength{\tabcolsep}{1em}
    \centering
    \begin{tabular}{l l} 
    \toprule
    \textbf{Attribute} & \textbf{\#Clusters ($k_{attr}$)} \\
    \midrule
     Speed &  14 \\
     Light Status &  2 \\
     Waypoints &  48 (24 per dim.) \\
     Target Points &  32 (16 per dim.) \\
    \bottomrule
    \end{tabular}
    \caption{\textbf{Number of Clusters used in K-means.} We use separate clusters for each dimension in the case of multi-dimensional attributes. Although light status is binary, we use k-means with 2 clusters for consistency across attributes.}
    \label{tab:kmeansparams}
\end{table}

\boldparagraph{Quantization}
We quantize the traffic light status, speed, waypoints, and target points using k-means to assign them to the cluster index. For waypoints and target points, which are both 2-dimensional, we apply k-means clustering on every dimension separately and quantize them into two indices in a similar manner. We list the number of clusters for each attribute in \tabref{tab:kmeansparams}.

\begin{table}[t]
    \setlength{\tabcolsep}{1em}
    \centering
    \begin{tabular}{l l} 
    \toprule
    \textbf{Hyper-param} & \textbf{Value} \\
    \midrule
     Hidden Dim. &  768 \\
     MLP Dim. & 3072 \\
     Attention Heads &  12 \\
     Number of Layers &  6 \\
     Weight Initialization &  distilgpt2~\cite{Sanh2019EMC2} \\
    \bottomrule
    \end{tabular}
    \caption{\textbf{Hyper-parameters of the GPT-2 Backbone.} We initialize the weights using the publicly available checkpoint of distilgpt2~\cite{Sanh2019EMC2}.}
    \label{tab:gpt2params}
\end{table}

\boldparagraph{GPT Backbone}
We use a slightly modified version of GPT-2 \cite{Radford2019} as the backbone. We use the Transformers library~\cite{Wolf2020ACL} and modify the embedding layer and the attention layers to implement block attention. We list the hyper-parameters for the GPT-2 model used in \tabref{tab:gpt2params}.

\section{Training Details}

\begin{table}[h]
    \setlength{\tabcolsep}{1em}
    \centering
    \begin{tabular}{l l} 
    \toprule
    \textbf{Hyper-param} & \textbf{Value} \\
    \midrule
     Epochs &  100 \\
     Warmup Epochs &  5 \\
     Learning Rate &  5e-5 \\
     Optimizer &  AdamW \\
     Weight Decay &  1e-4 \\
     Effective Batch Size &  512 \\
     Grad. Norm Clip &  1.0 \\
    \bottomrule
    \end{tabular}
    \caption{\textbf{Training Hyper-parameters of CarFormer.}}
    \label{tab:hyperparams-carformer}
\end{table}

\label{sec:train_details}
We train our model for 100 epochs on the full dataset and pick the checkpoint with the best validation loss to evaluate. We use a setup of either $4\times$ A6000s or $4\times$ A100s to train our models. We keep the effective batch size as 512 across runs. Moreover, we weigh the two loss terms by multiplying the forecasting loss by 40 as shown in \eqref{eq:loss_weights}. We list the other hyper-parameters in \tabref{tab:hyperparams-carformer}.

\begin{equation}
    \label{eq:loss_weights}
    \cL = 40~\cL_{\textit{forecast}} + \cL_{\textit{wp}}
\end{equation}

\begin{table*}[t]
\setlength{\tabcolsep}{0.75em}
\centering
\begin{tabular}{l l l l} 
\toprule
 & \textbf{SAVi-30-light} & \textbf{SAVi-7-light} & \textbf{SAVi-7-base} \\
 \midrule
 Effective Batch Size &  256 &  256 &  256  \\
 Training Steps &  80000 &  80000 &  80000  \\
 Optimizer &  Adam &  Adam &  Adam  \\
 Learning Rate &  1e-4 &  1e-4 &  1e-4  \\
 Gradient Clip &  5e-2 &  5e-2 &  5e-2  \\
 Warmup steps &  4000 &  4000 &  4000  \\
 Context Length &  2 frames &  2 frames &  2 frames  \\
 Data FPS & 2 fps & 2 fps & 2 fps  \\
 \textbf{Num. Slots} &  \textbf{30} & \textbf{7} & \textbf{7} \\
 Slot Size &  128 &  128 &  128  \\
 Slot MLP size &  256 &  256 &  256  \\
 Slot Attn. Iterations &  2 &  2 &  2  \\
 Enc. Filter Sizes & (64, 64, 64, 64) & (64, 64, 64, 64) & (64, 64, 64, 64)\\
 Enc. Kernel Sizes & (5, 5, 5, 5) & (5, 5, 5, 5) & (5, 5, 5, 5) \\
 \textbf{Enc. Strides} & \textbf{(2, 1, 1, 1)} & \textbf{(2, 1, 1, 1)} & \textbf{(1, 1, 1, 1)}  \\
 \textbf{Enc. MLP dim.} & \textbf{128} & \textbf{128} & \textbf{256} \\
 Dec. Init. Resolution & (128, 24, 24) & (128, 24, 24) & (128, 24, 24) \\
 \textbf{Dec. Filter Sizes} & \textbf{(64, 32, 16, 8)} & \textbf{(64, 32, 16, 8)} & \textbf{(64, 64, 64, 64)} \\
 Dec. Kernel Sizes & (5, 5, 5, 5) & (5, 5, 5, 5) & (5, 5, 5, 5) \\
 Dec. Strides & (2, 2, 2, 1) & (2, 2, 2, 1) & (2, 2, 2, 1)  \\
 Pred. Type &  Transformer &  Transformer &  Transformer  \\
 Pred. Layers &  2 &  2 &  2  \\
 Pred. MLP Size &  512 &  512 &  512  \\
 Pred. Attn. Heads &  4 &  4 &  4  \\
\bottomrule
\end{tabular}
\caption{\textbf{Hyper-parameters of SAVi.} We refer to the version of the encoder with a light decoder as SAVi-30-light and SAVI-7-light. We refer to the original SAVi model as SAVi-7-base. We show the hyper-parameters that are different across models in bold.}
\label{tab:saviparams}
\end{table*}

\section{Additional Quantitative Results}
\label{sec:add_quan}

\boldparagraph{Ablation of Light SAVi}
We modify the architecture of SAVi to be able to increase the number of slots while keeping slot extraction feasible. Specifically, we propose a lighter decoder by decreasing the number of parameters. To measure the effect of these changes quantitatively, we train our model with the light decoder while keeping the number of slots constant at 7. Note that, training the slot extraction model is prohibitively slow with the original decoder when we increase the number of slots to 30. 

\begin{table}[ht]
   \setlength{\tabcolsep}{0.75em}
   \centering
   \begin{tabular}{l c c c c c c} 
   \toprule
   \textbf{Method} & \textbf{\#Slots} & \textbf{LD} &  \textbf{\#Steps} & \textbf{ARI$\uparrow$} & \textbf{mIoU$\uparrow$} \\ 
   \midrule
   \multirow{3}{*}{SAVi}  & 
          7  & \xmark & - &  0.905  &  0.856 \\ %
        & 7  & \cmark & - &  0.911  & 0.860 \\
        & 30 & \cmark & - &  0.924  &  0.874 \\ %
   \midrule
   \multirow{6}{*}{CarFormer} & 
     7  & \xmark & 1 &  0.742  &  0.644 \\ %
   & 7  & \cmark & 1 &  0.704  &  0.602 \\ %
  & 30  & \cmark & 1 &  0.795   &  0.702 \\ %
   & 7  & \xmark & 4 &  0.513  &  0.436 \\ %
   & 7  & \cmark & 4 &  0.462 &  0.385 \\ %
   & 30 & \cmark & 4 &  0.540  &  0.454 \\ %
      \midrule
   \multirow{2}{*}{Input-Copy} & 
     - & - & 1 &  0.641  &  0.561 \\ %
   & - & - & 4 &  0.412  &  0.375 \\ %
   \bottomrule
   \end{tabular}
   \caption{\textbf{Forecasting Results with Light Decoder (LD).} We show the results for SAVi reconstruction, which serve as the upper bound as the model has access to ground truth. We compare six versions of our model, with each of the three SAVi versions (7-base, 7-light, 30-light), and trained to predict 1 or 4 timesteps into the future. Finally, we show the results of predicting the current BEV to forecast the future as a baseline.} 
   \label{tab:forecast-supp-lightdec}
   \end{table}

We evaluate the forecasting performance of the model in \tabref{tab:forecast-supp-lightdec}. Although the performance degrades slightly with the light decoder in the case of 7 slots, increasing the number of slots to 30 improves performance noticeably. Increasing the number of slots is not only crucial to improve the performance in terms of slot extraction, it also proves to be crucial in terms of driving performance as we show in the main paper (\tabref{tab:slots}).
To further evaluate its effect on driving performance, we compare CarFormer using the original SAVi checkpoint with 7 slots (full decoder) as well as the light decoder version with 7, 14, and 30 slots in \tabref{tab:slots-light-ablation}. The results show that increasing the number of slots improves the driving performance and the best driving performance is achieved with 30 slots which was only made possible with the light decoder. 

\begin{table}[ht]
\setlength{\tabcolsep}{0.75em}
\centering
\begin{tabular}{c c c c c} 
\toprule
\textbf{LD} & \textbf{\#Slots} & \textbf{DS$\uparrow$} & \textbf{IS$\uparrow$} & \textbf{RC$\uparrow$}\\ 
\midrule
\xmark & 7 & 60.82{\small$\pm$0.87} & 0.73{\small$\pm$0.02} & 79.44{\small$\pm$1.62} \\  %

\cmark & 7 & 62.93{\small$\pm$6.78} & 0.73{\small$\pm$0.07} & 80.20{\small$\pm$1.87} \\ %
\cmark & 14 & 69.75{\small$\pm$8.03} & 0.78{\small$\pm$0.04} & 86.17{\small$\pm$2.79} \\ 
\cmark & 30 & 74.89{\small$\pm$1.44} & 0.79{\small$\pm$0.02} & 92.90{\small$\pm$1.28} \\
\bottomrule
\end{tabular}
\caption{\textbf{Driving Performance with Light Decoder (LD).} We test three different SAVi versions. Light refers to the lighter version of SAVi we propose, and the base version otherwise. \#Slots refers to the number of slots used in the model. We report mean$\pm$std of 3 different runs on the Longest6.}
\label{tab:slots-light-ablation}
\end{table}

\boldparagraph{The Effect of Enlarging Slots on Forecasting}
Slot extraction is particularly difficult for small objects due to the nature of the reconstruction loss being dominated by larger objects. As a simple fix, we enlarge small vehicles in the BEV input before slot extraction. In particular, we enlarge any vehicle that is smaller than $4.9~m$ in length or $2.12~m$ in width to a length of $4.9~m$ and a width of $2.12~m$, respectively. We set the dimensions according to Lincoln MKZ, the default ego vehicle on CARLA.

\begin{table}[t]
   \setlength{\tabcolsep}{0.75em}
   \centering
   \begin{tabular}{l c c c c c} 
   \toprule
   \textbf{Method} & \textbf{\#Slots} & \textbf{Enlarge} &  \textbf{\#Steps} & \textbf{ARI$\uparrow$} & \textbf{mIoU$\uparrow$} \\ 
   \midrule
   \multirow{2}{*}{SAVi}  & 
   \multirow{2}{*}{7} 
      & \xmark & - &  0.900  &  0.846 \\ %
   &  & \cmark & - &  0.905  &  0.856 \\ %
   \addlinespace
   \hdashline
   \addlinespace
   \multirow{4}{*}{CarFormer} & 
   \multirow{4}{*}{7} 
      & \xmark & 1 &  0.712  &  0.611 \\ %
   &  & \cmark & 1 &  0.742  &  0.644 \\ %
   &  & \xmark & 4 &  0.487  &  0.412 \\ %
   &  & \cmark & 4 &  0.513  &  0.436 \\ %
   \addlinespace
   \midrule
   \addlinespace
   \multirow{2}{*}{SAVi}  & 
   \multirow{2}{*}{30}  
            & \xmark & - &  0.910  &  0.850 \\
      &  & \cmark & - &  0.924  &  0.873 \\ 
   \addlinespace
   \hdashline
   \addlinespace
   \multirow{4}{*}{CarFormer} 
   & \multirow{4}{*}{30} 
         & \xmark & 1 &  0.765  &  0.664 \\ 
      &  & \cmark & 1 &  0.795  &  0.703 \\ 
      &  & \xmark & 4 &  0.508  &  0.422 \\ 
      &  & \cmark & 4 &  0.540  &  0.454 \\ 
   \bottomrule
   \end{tabular}
   \caption{\textbf{Forecasting Results by Enlarging Slots.} We compare the forecasting performance of the base SAVi model for the 7-slot case (SAVi-7-base) and the light version for the 30-slot case (SAVi-30-light) for predicting 1 or 4 steps ahead.} 
   \label{tab:enlarging_ablation}
\end{table}

We evaluated the effect of enlarging slots on the driving performance in the main paper (\tabref{tab:slots}). Here, we investigate its effect on forecasting. We compare two models, one with the base SAVi with 7 slots (SAVi-7-base), and our version of SAVi with the light decoder and 30 slots (SAVi-30-light) in \tabref{tab:enlarging_ablation}. Enlarging small vehicles consistently improves forecasting performance in all settings and in terms of all metrics.

\boldparagraph{The Effect of Block Attention on Slot Forecasting}
In \tabref{tab:ablation} of the main paper, we show that block attention (BA) between slots improves driving performance, and hypothesize that this allows us to better model the relationship between objects. To further validate the role of BA in relating objects, we evaluate the forecasting performance with and without BA in \tabref{tab:blockattn-forecasting-supp}. The model with BA results in consistent improvements in forecasting performance with both 14 and 30 slots, 
which shows its positive effect by allowing more interaction between objects in the autoregressive model compared to regular causal attention. We outline how we can retain the ability to autoregressively rollout future time steps in \secref{sec:rollout_gen}. 

\begin{table}[ht]
   \setlength{\tabcolsep}{0.75em}
   \centering
   \begin{tabular}{c c l l} 
   \toprule
   \textbf{BA} &  \textbf{\#Slots} & \textbf{ARI$\uparrow$} & \textbf{mIoU$\uparrow$} \\ 
   \midrule
           \xmark & 14 &  0.511  &  0.432 \\ %
           \cmark & 14 &  0.533~\small{(+0.022)}  &  0.449~\small{(+0.017)} \\ %
           \midrule
           \xmark & 30 &  0.509  &  0.428 \\ %
           \cmark & 30 &  0.540~\small{(+0.031)}  &  0.454~\small{(+0.026)} \\ %
   \bottomrule
   \end{tabular}
   \caption{\textbf{Forecasting Results with Block Attention (BA).} We evaluate CarFormer with our optimized SAVi with either 14 or 30 slots as the slot extractor backbone. The model with BA results in consistent improvements in all forecasting metrics.}
   \label{tab:blockattn-forecasting-supp}
   \end{table}

\boldparagraph{Ablating the Design of Block Attention}
In our main model, we place the object level features, whether in the form of slots or as attributes, in the same block as the desired route $\br_t^1, \br_t^2$. We ablate different possible options for the block in \tabref{tab:blkattndesign-ablation}. Although removing the desired route from the block leads to slight performance gains in DS, we include the desired route in the block due to a lower standard deviation in driving score.

\begin{table*}[t]
\setlength{\tabcolsep}{0.75em}
\centering
\begin{tabular}{c c c c c | c c c} 
\toprule
\multicolumn{5}{c}{Input Feature} & & & \\
\textbf{TP} & \textbf{S} & \textbf{TL} & \textbf{O} & \textbf{R} &\textbf{DS$\uparrow$} & \textbf{IS$\uparrow$} & \textbf{RC$\uparrow$} \\
\midrule
\xmark & \xmark & \xmark & \cmark & \cmark & 74.89{\small$\pm$1.44} & 0.79{\small$\pm$0.02} & 92.90{\small$\pm$1.28} \\
\xmark & \xmark & \xmark & \cmark & \xmark & 75.06{\small$\pm$2.10} & 0.80{\small$\pm$0.01} & 90.99{\small$\pm$1.26} \\
\xmark & \cmark & \cmark & \cmark & \xmark & 70.01{\small$\pm$2.79} & 0.79{\small$\pm$0.02} & 86.23{\small$\pm$1.71} \\
\cmark & \cmark & \cmark & \cmark & \cmark & 72.34{\small$\pm$2.53} & 0.78{\small$\pm$0.02} & 91.21{\small$\pm$1.07} \\
\bottomrule
\end{tabular}
\caption{\textbf{Block attention design ablation.} We compare placing different inputs into the attention block. Since our design contains only one block, a tick mark means the feature is included in the block.  Results are of 3 runs on Longest6. TP: Target Point, S: Speed, TL: Traffic Light, O: Objects (attributes/slot features), R: Desired route.} 
\label{tab:blkattndesign-ablation}
\end{table*}

\boldparagraph{Ablation of Target Point Input}
In our approach, we feed the next target point in the desired route as an input to the backbone. With the desired route being also given as input, this could possibly be redundant. We compare the results with and without the target point in \tabref{tab:tp-ablation}. The inclusion of the next target point improves the DS, IS, and the RC of the agent.

\begin{table*}[b]
\setlength{\tabcolsep}{0.75em}
\centering
\begin{tabular}{c | c c c} 
\toprule
\textbf{TP} & \textbf{DS$\uparrow$} & \textbf{IS$\uparrow$} & \textbf{RC$\uparrow$} \\
\midrule
\cmark & 74.89{\small$\pm$1.44} & 0.79{\small$\pm$0.02} & 92.90{\small$\pm$1.28} \\
\xmark & 72.34{\small$\pm$0.35} & 0.77{\small$\pm$0.00} & 90.70{\small$\pm$0.66} \\
\bottomrule
\end{tabular}
\caption{\textbf{Difference with versus without target point as input} We compare CarFormer with and without using target points as input. Results are of 3 runs on Longest6.} 
\label{tab:tp-ablation}
\end{table*}

\boldparagraph{Increasing the Input Context Size} In all the experiments in the main paper, the model only sees the information from the current time step. Although BEV from previous time steps is used within SAVi, we only use the slot features of the current time step. To explore the effectiveness of extending our model to multiple time steps, we train the model using a context length of up to 4 and evaluate the forecasting performance in \tabref{tab:multi_context_forecast}. We observe consistent gains in both forecasting metrics as we increase the context length, showing the model is able to utilize information from previous timesteps to guide its future prediction.

\begin{table}[h]
\centering
\begin{tabular}{c c c c c c c c} 
\toprule
\multicolumn{8}{c}{\textbf{Context Length}} \\
\multicolumn{2}{c}{\textbf{1}} & &
\multicolumn{2}{c}{\textbf{2}} & & \multicolumn{2}{c}{\textbf{4}} \\ 
\cline{1-2} \cline{4-5} \cline{7-8}
 \textbf{ARI} & \textbf{mIoU} & & \textbf{ARI} & \textbf{mIoU} & & \textbf{ARI} & \textbf{mIoU}  \\ 
\midrule
0.795 & 0.702 & & 0.816 & 0.728 & & 0.824  &  0.740  \\
\bottomrule
\end{tabular}
\caption{\textbf{Effect of input context length on forecasting:} Increasing the context length consistently improves forecasting performance in all metrics, showing that the model is able to use the information from previous timesteps to better predict the future.}
\label{tab:multi_context_forecast}
\end{table}

\boldparagraph{Effect of Forecasting Steps}
To further investigate the effect of forecasting, we experiment with the number of steps into the future that we forecast or $f$ in \eqref{eq:loss_forecast_scene}. As can be seen in \tabref{tab:frcstepablation}, increasing the number of forecasting steps results in improvements in both infraction score and route completion, and consequently a higher driving score. As a result, unless otherwise specified, all our models have been trained to predict timestep $t+4$.

\begin{table*}[t]
\setlength{\tabcolsep}{0.75em}
\centering
\begin{tabular}{c c c c} 
\toprule
\textbf{\#Steps} & \textbf{DS$\uparrow$} & \textbf{IS$\uparrow$} & \textbf{RC$\uparrow$} \\
\midrule
1 & 71.40{\small$\pm$0.74} & 0.78{\small$\pm$0.01} & 89.25{\small$\pm$0.99} \\
4 & 74.89{\small$\pm$1.44} & 0.79{\small$\pm$0.02} & 92.90{\small$\pm$1.28} \\
\bottomrule
\end{tabular}
\caption{\textbf{Varying Forecasting Time steps.} We compare varying the number of timesteps into the future that we forecast.} 
\label{tab:frcstepablation}
\end{table*}

\begin{figure*}[t]
    \centering
        \includegraphics[width=.49\linewidth]{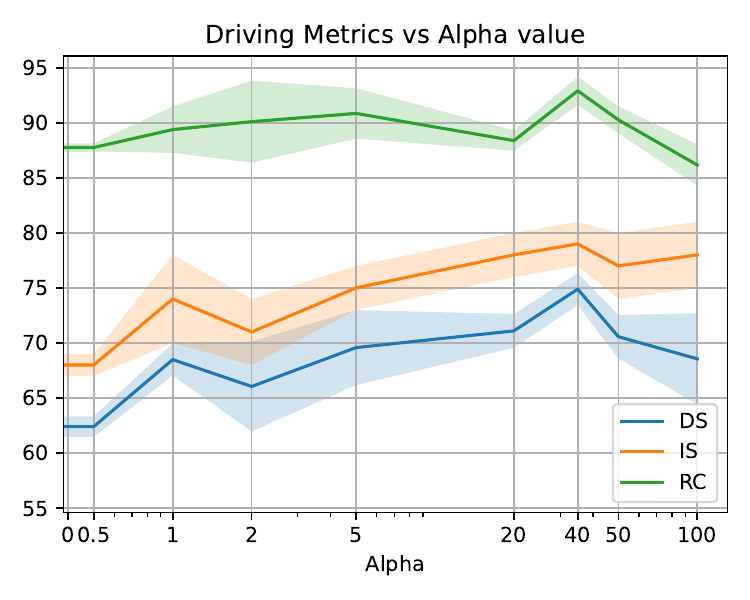}
        \label{fig:alphaablationfig}
    \caption{\textbf{Ablation of Forecasting Weight.} We visualize the effect of varying the hyper-parameter $\alpha$, the weight of forecasting in the loss function, on driving performance. While initially increasing the contribution of forecasting improves the driving performance, it peaks around 40. We use CarFormer with SAVi-30-light as the encoder backbone for all experiments.}
    \label{fig:alphaablation}
\end{figure*}

\boldparagraph{Effect of Forecasting Weight}
We evaluate the effect of varying the weight of the forecasting term, $\alpha$ in the loss function \eqref{eq:total_loss} on driving performance in \figref{fig:alphaablation}. While initially increasing the importance of forecasting improves the driving performance, the performance peaks around $40$ and drops for larger values of $\alpha$. This could be because the training signal from forecasting eventually dominates the parameter updates to the model, reducing the relative importance of the actual driving task.

\boldparagraph{Efficiency}
In \tabref{tab:inference-time}, we list the inference time in milliseconds (ms) and the number of parameters in millions (M), trainable parameters in parenthesis, for different versions of our model compared to PlanT. We measure the inference time for all models on a machine with i9-10900K CPU and RTX 3090 GPU. For our models, we report the results with object attributes similar to PlanT, with VQ-VAE, and with 7, 14, and 30 slots. All our models can run in real-time, with over 60 FPS. We also report the results without SAVi to show that around half of the time is spent on slot extraction. The inference time can be significantly improved with more efficient slot extraction techniques. 

\begin{table}[ht]
\setlength{\tabcolsep}{0.75em}
\centering
\begin{tabular}{l l l} 
\toprule 
\textbf{Model} & \textbf{Time (ms)} & \textbf{\#Params} \\ 
\midrule 
PlanT & 6.41 & 42M \\ 
\midrule
Obj. Attributes & 7.31 & 53M \\ 
VQ-VAE & 8.51 & 60M~(43M) \\ 
7 Slots & 13.87 & 54M~(44M) \\ 
14 Slots & 13.94 & 54M~(44M) \\ 
30 Slots & 14.02 & 54M~(44M) \\ 
\midrule
7 Slots w/o SAVi & 7.46 & 54M~(44M) \\ 
14 Slots w/o SAVi & 7.37 & 54M~(44M) \\ 
30 Slots w/o SAVi & 7.44 & 54M~(44M) \\ 
\bottomrule    
\end{tabular}
\caption{\textbf{Inference Time and Number of Parameters.} We show inference time (in ms) and the number of parameters (in M) for different versions of our model compared to PlanT Medium \cite{Chitta2023PAMI}. When using slots as well as a VQ-VAE, we use a frozen backbone. In addition to the number of parameters, we show the number of trainable parameters in parenthesis.} 
\label{tab:inference-time}
\end{table}

\section{Additional Qualitative Results}
\label{sec:add_qual}

\subsection{Effect of Increasing the Number of Slots}
\begin{figure*}[t]
    \centering
    \begin{minipage}{0.49\textwidth}
        \subfloat[]{\includegraphics[width=.99\linewidth]{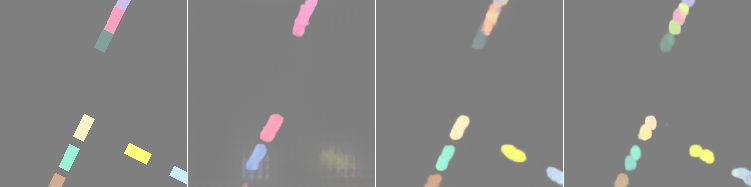} \label{fig:fsza}} \\ %
        \subfloat[]{\includegraphics[width=.99\linewidth]{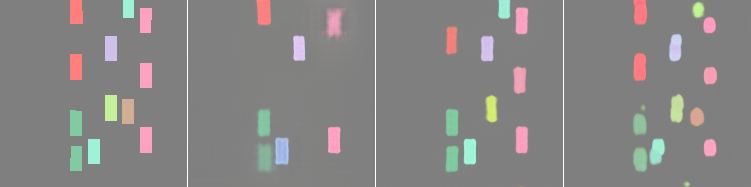}\label{fig:fszb}} \\ %
        \subfloat[]{\includegraphics[width=.99\linewidth]{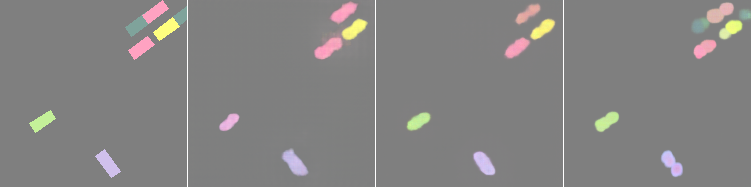}\label{fig:fszc}} \\ %
    \end{minipage}%
    \hfill
    \begin{minipage}{0.49\textwidth}
        \subfloat[]{\includegraphics[width=.99\linewidth]{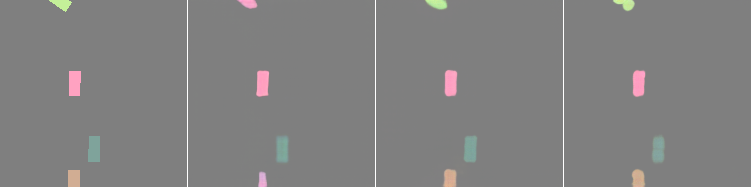}\label{fig:fszd}} \\ %
        \subfloat[]{\includegraphics[width=.99\linewidth]{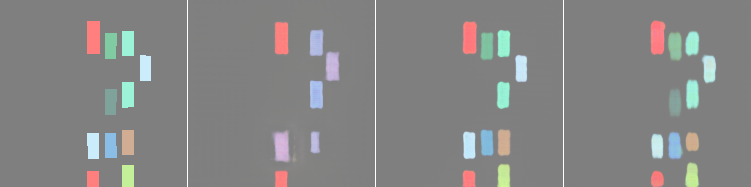}\label{fig:fsze}} \\ %
        \subfloat[]{\includegraphics[width=.99\linewidth]{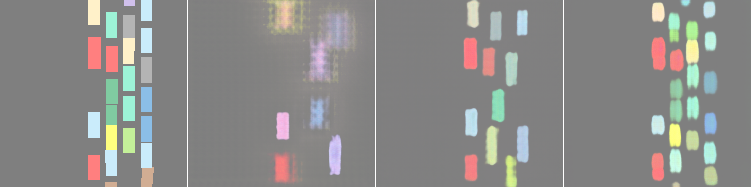}\label{fig:fszf}} \\ %
    \end{minipage}%
    \caption{\textbf{Comparison of Slot Extraction by Varying the Number of Slots.} Within each sub-figure, the columns correspond to ground truth BEV followed by SAVi-light reconstructions with 7, 14, and 30 slots, respectively.}
    \label{fig:slotsize}
\end{figure*}

We visually compare the reconstructions of the different versions of SAVi to better understand the effect of increasing the number of slots. 
As can be seen in \figref{fig:slotsize}, the SAVi model we use with 7 slots cannot capture all the vehicles in the scene. Some vehicles are either missing completely as in \subref{fig:fszb}, \subref{fig:fsze}, or replaced with blurry regions as in  \subref{fig:fsza}, \subref{fig:fszf}. The SAVi model with 14 slots faces the same issue, albeit to a lesser extent. For example, there is a missing vehicle in the top right of \subref{fig:fszb}, as well as multiple missing vehicles in \subref{fig:fsze} and \subref{fig:fszf}, including one very close to the ego vehicle in \subref{fig:fsze}, which could result in a collision.

When increasing the number of slots to 30, the model can capture all vehicles to some extent. 
One side effect of increasing the number of slots to 30 is that multiple slots can bind to a single vehicle, which can be seen in the reconstructions. For example, the vehicles in the bottom of \subref{fig:fszc} appear to be reconstructed as multiple blobs. However, the model can capture all vehicles in the scene, which empirically improved driving performance and infraction scores.

\subsection{Additional Slot Forecasting Visualizations}
\begin{figure*}[t]
    \centering
    \begin{minipage}{0.49\textwidth}
        \subfloat[]{\includegraphics[width=.99\linewidth]{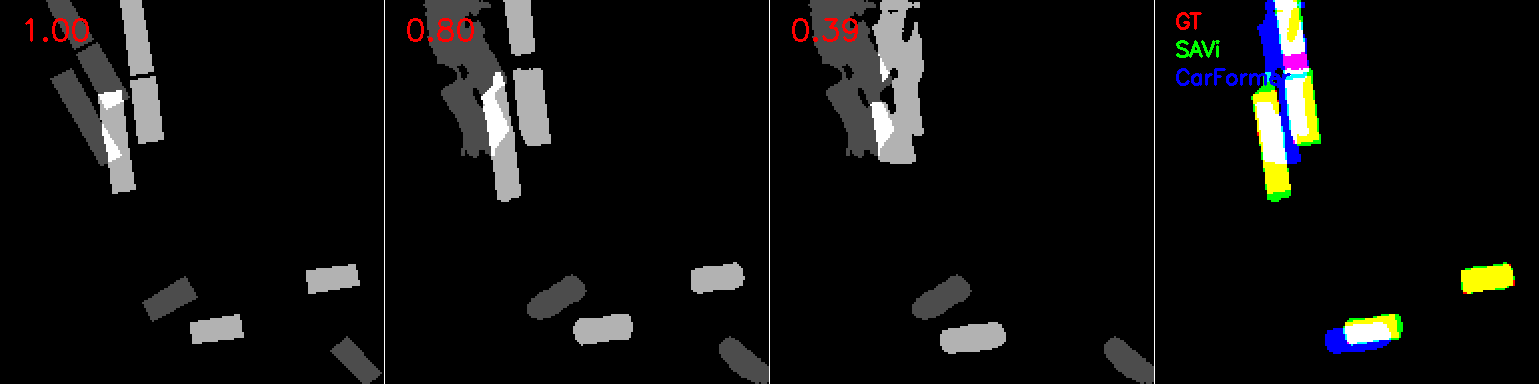} \label{fig:ffca}} \\ %
        \subfloat[]{\includegraphics[width=.99\linewidth]{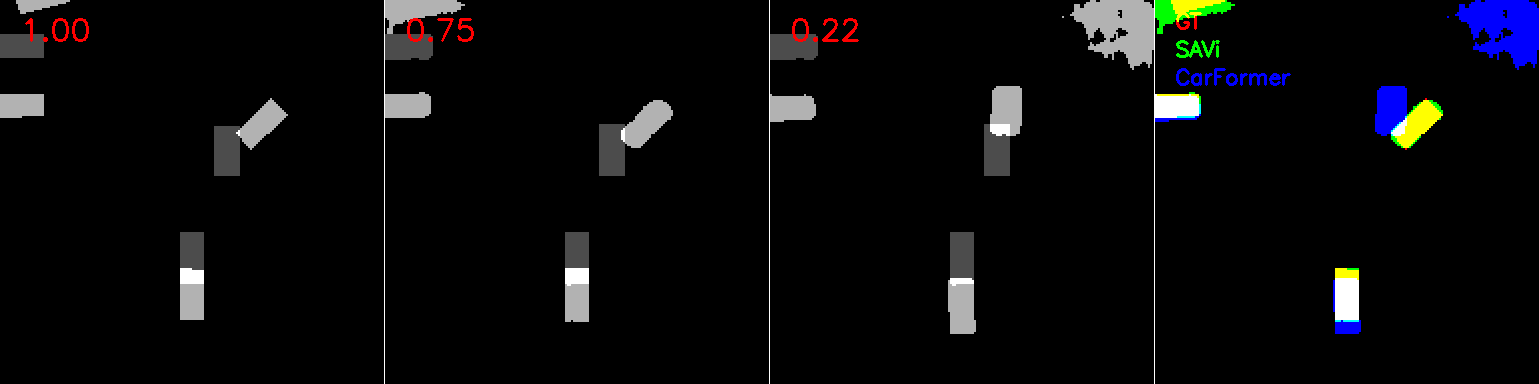} \label{fig:ffcb}} \\ %
        \subfloat[]{\includegraphics[width=.99\linewidth]{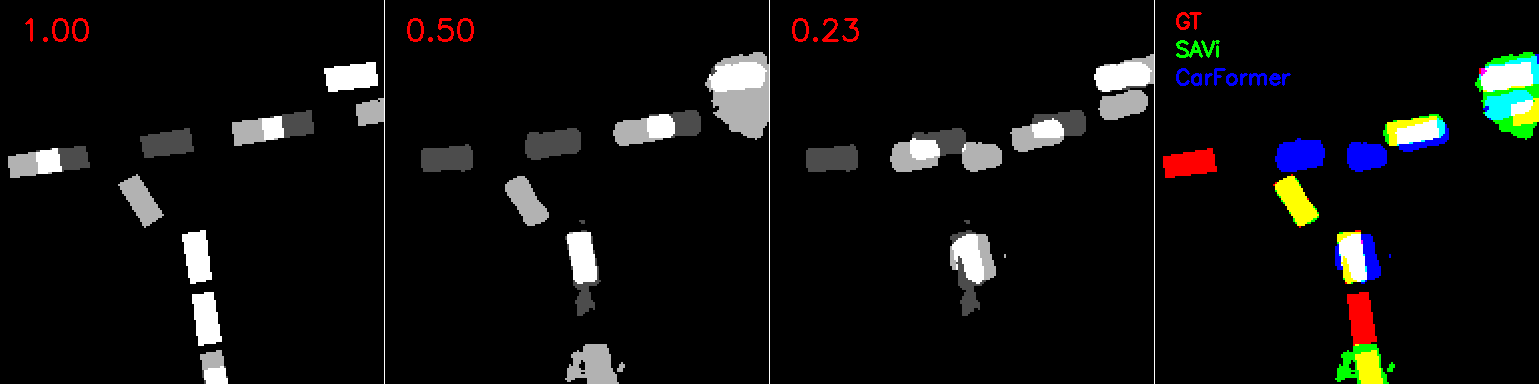} \label{fig:ffcc}} \\ %
        \subfloat[]{\includegraphics[width=.99\linewidth]{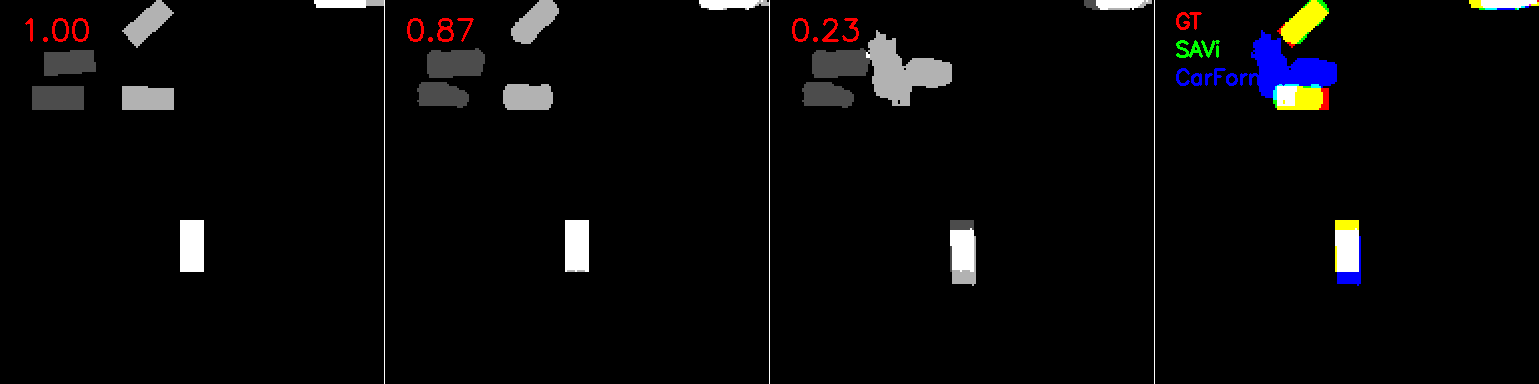} \label{fig:ffcd}} \\ %
        \subfloat[]{\includegraphics[width=.99\linewidth]{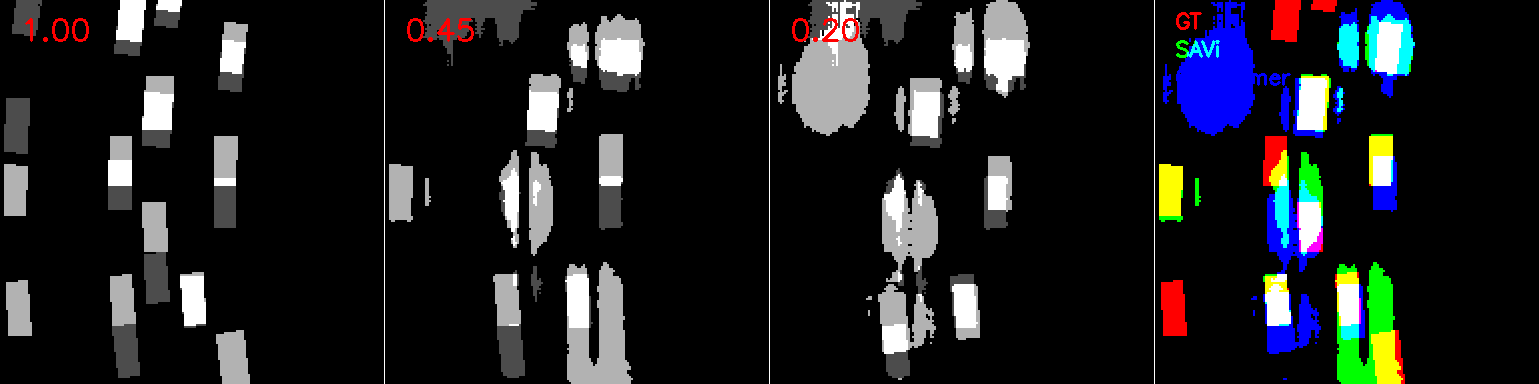} \label{fig:ffce}} \\ %
        \subfloat[]{\includegraphics[width=.99\linewidth]{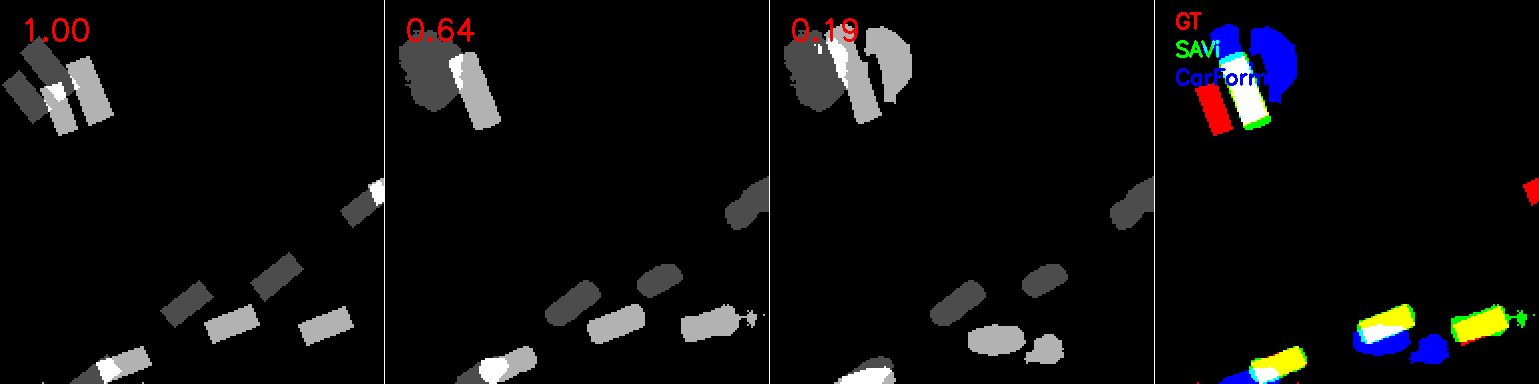} \label{fig:ffcf}} \\ %
        \subfloat[]{\includegraphics[width=.99\linewidth]{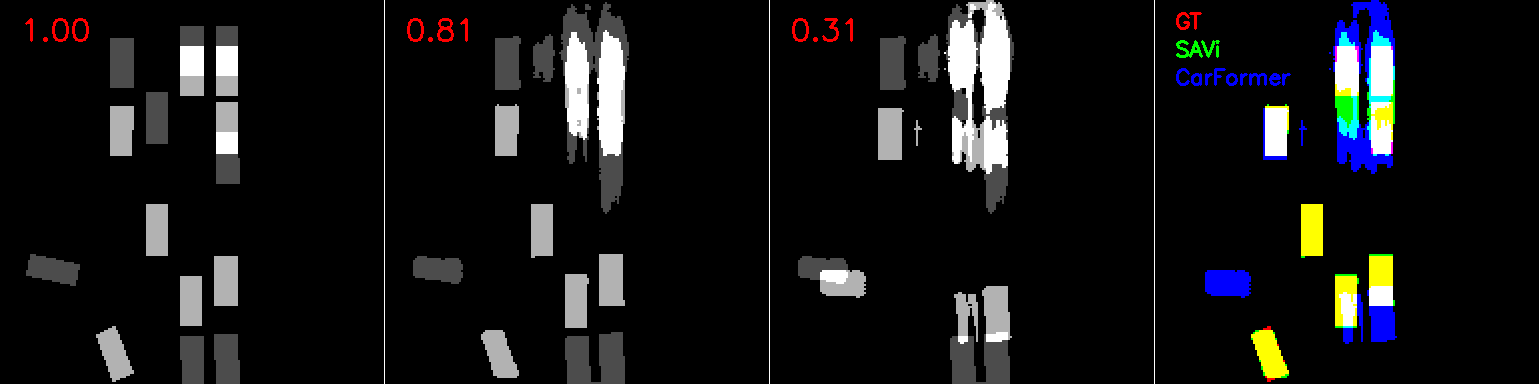} \label{fig:ffcg}} \\ %
    \end{minipage}%
    \hfill
    \begin{minipage}{0.49\textwidth}
        \subfloat[]{\includegraphics[width=.99\linewidth]{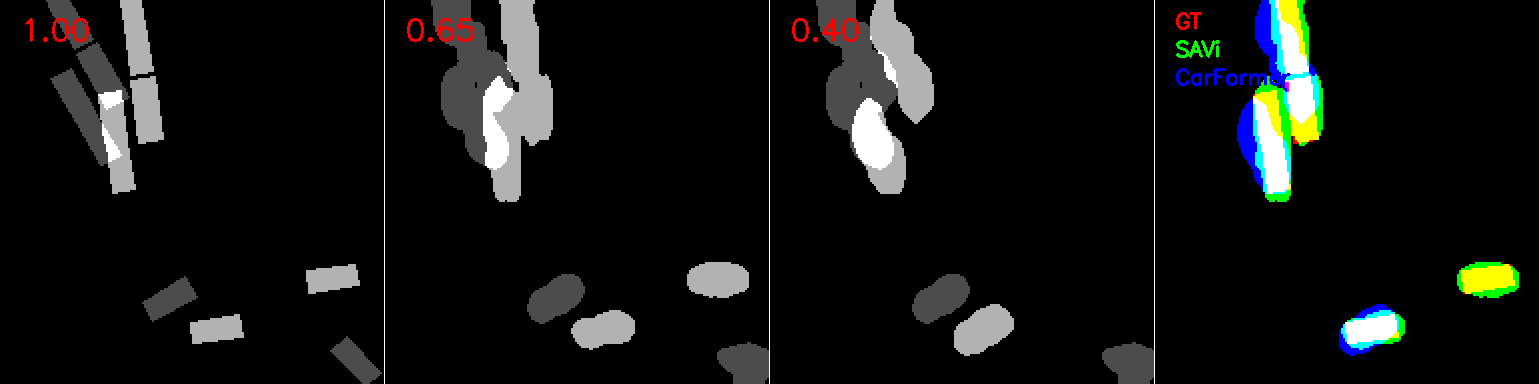} \label{fig:ffch}} \\ %
        \subfloat[]{\includegraphics[width=.99\linewidth]{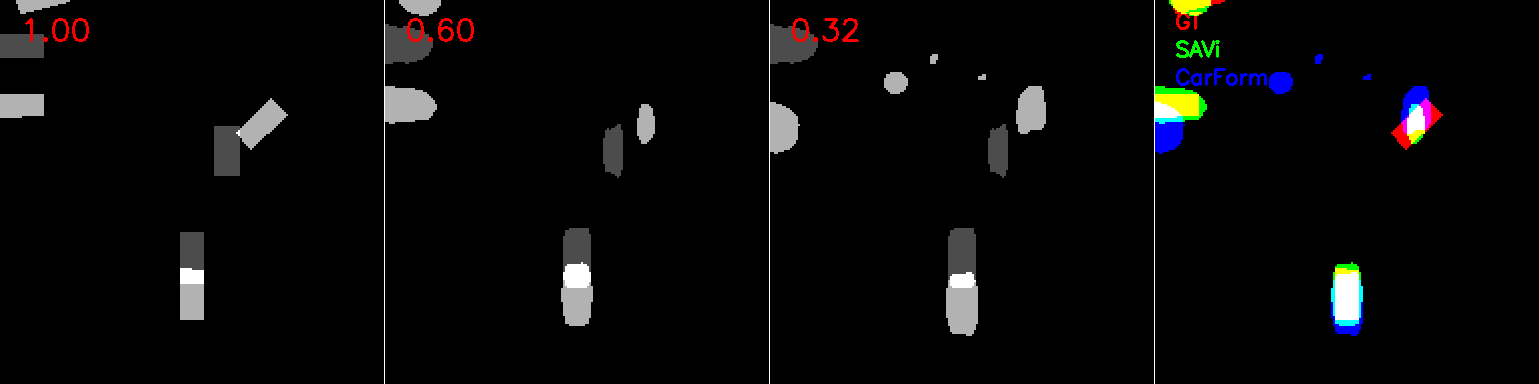} \label{fig:ffci}} \\ %
        \subfloat[]{\includegraphics[width=.99\linewidth]{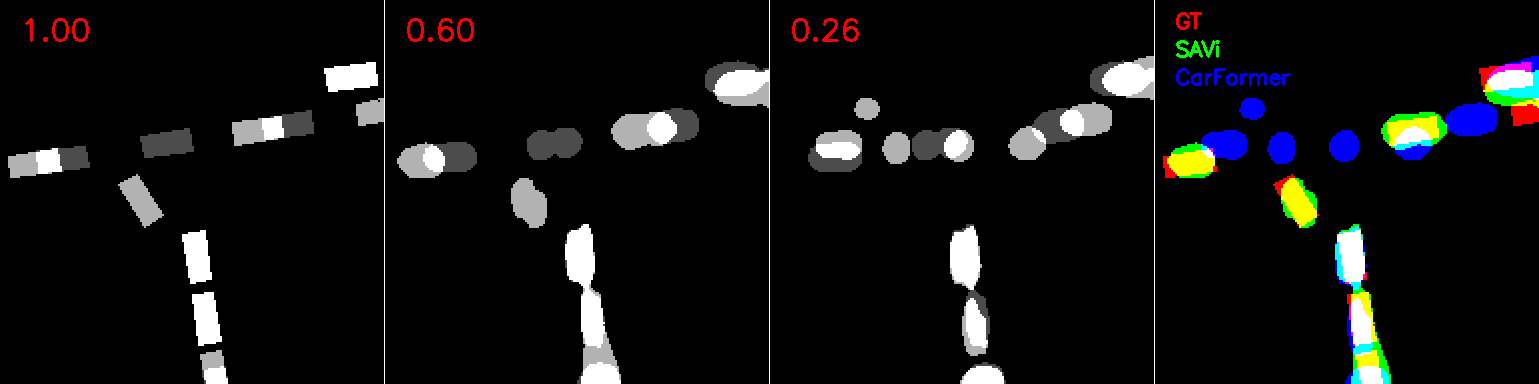} \label{fig:ffcj}} \\ %
        \subfloat[]{\includegraphics[width=.99\linewidth]{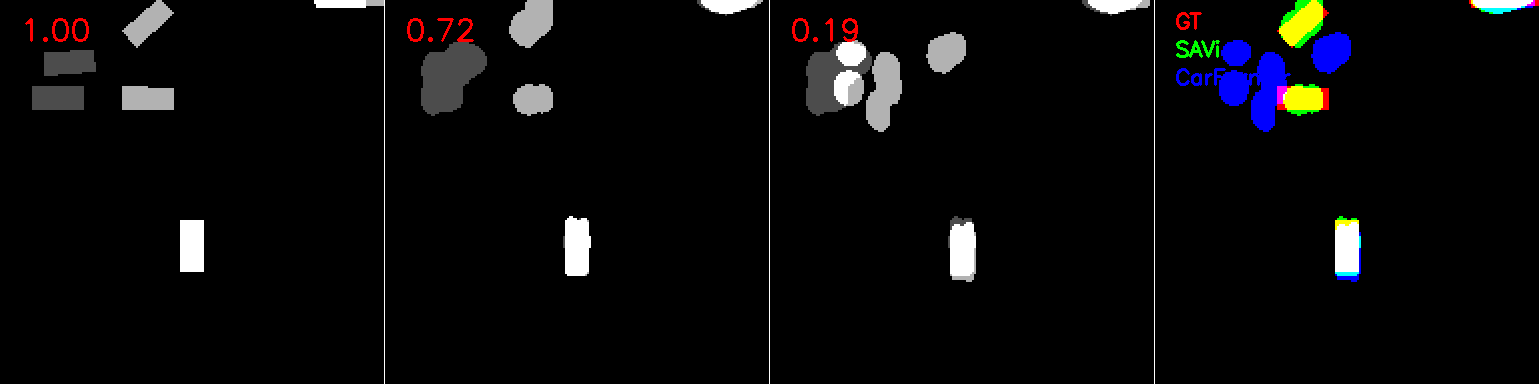} \label{fig:ffck}} \\ %
        \subfloat[]{\includegraphics[width=.99\linewidth]{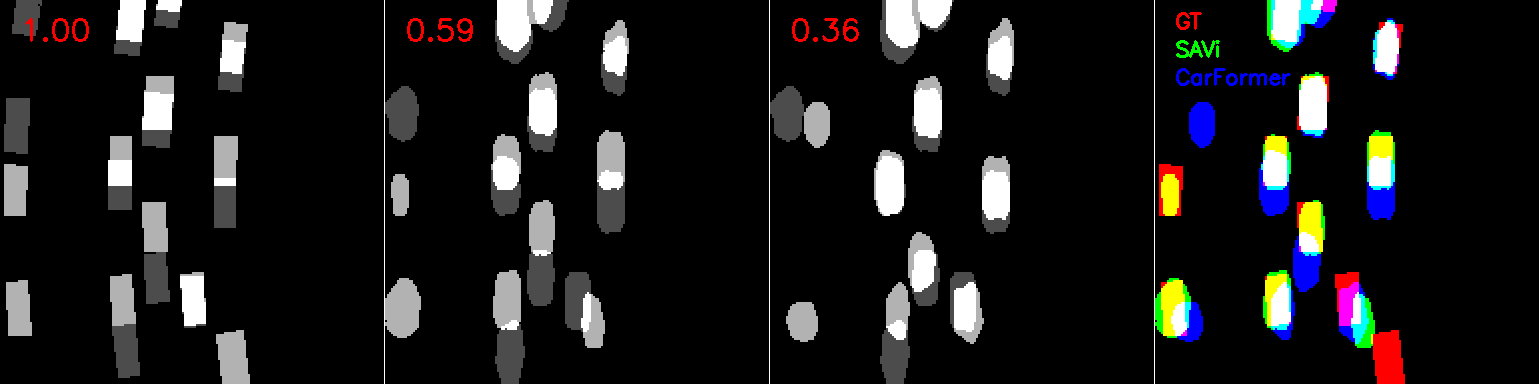} \label{fig:ffcl}} \\ %
        \subfloat[]{\includegraphics[width=.99\linewidth]{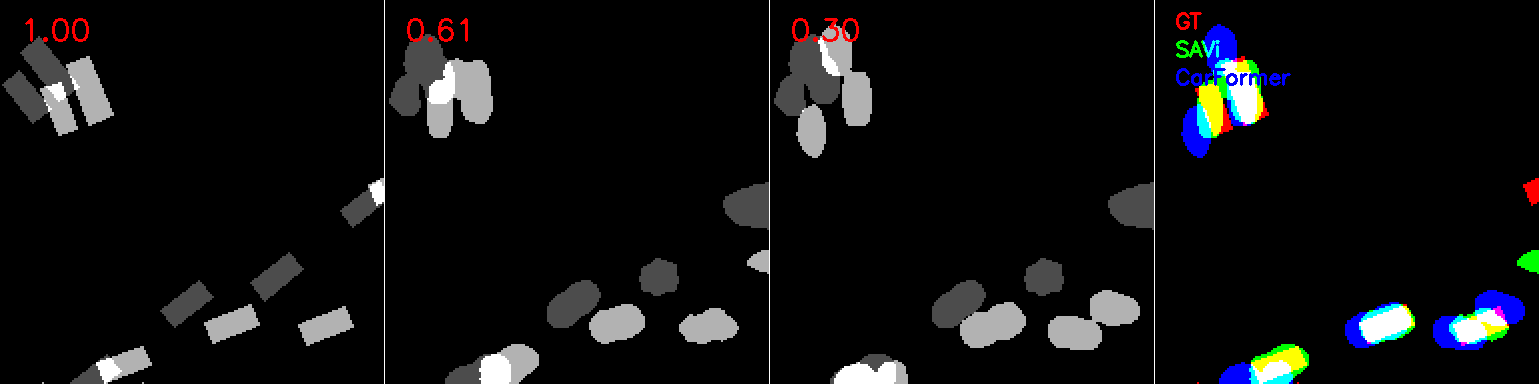} \label{fig:ffcn}} \\ %
        \subfloat[]{\includegraphics[width=.99\linewidth]{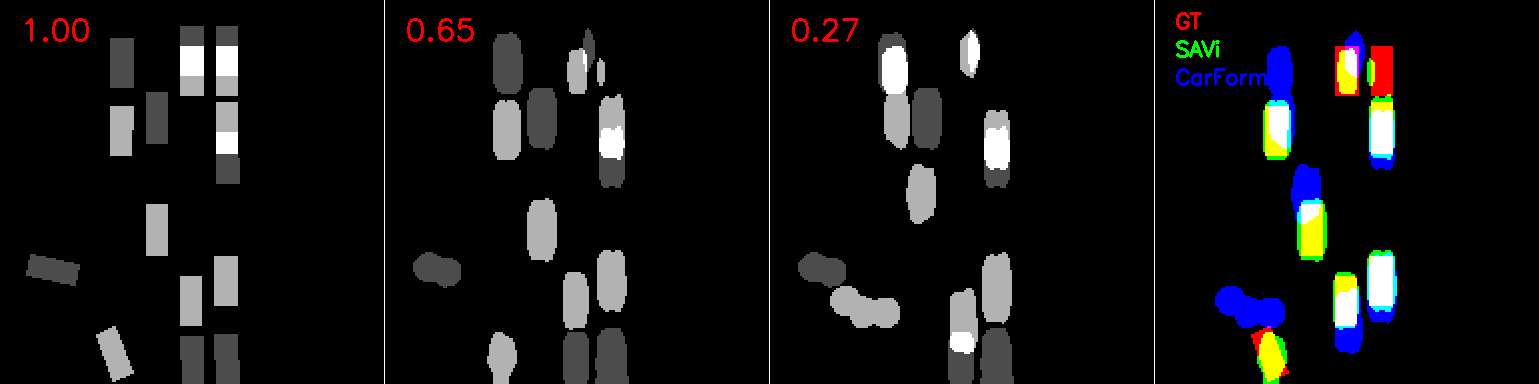} \label{fig:ffco}} \\ %
    \end{minipage}%
    \caption{\textbf{Additional Visualization of Slot Forecasting Results for CarFormer with SAVi-7-base (left half) and CarFormer with SAVi-14-light (right half).} Each sub-figure shows an example of input~(dark grey)-output~(light grey) objects in the first column, SAVi reconstructions in the second column, and our model's predictions in the third column. The top left corner of each column shows the mIoU compared to the ground truth. For comparison, we overlay the three in the last column where the red channel~(\textcolor{red}{R}) is the ground-truth location, the green channel~(\textcolor{green}{G}) is SAVi reconstruction, and the blue channel~(\textcolor{blue}{B}) is our prediction. In the case of perfect alignment between the three, we see the vehicles in white, and different errors can be seen from combinations of \textcolor{red}{R}-\textcolor{green}{G}-\textcolor{blue}{B} colors such as yellow~(\textcolor{yellow}{R+G}) indicating misses and blue indicating false positives for our model~(\textcolor{blue}{B}).}
    \label{fig:slotsforcsupp}
\end{figure*}

We provide additional samples of the model's future predictions in \figref{fig:slotsforcsupp}. In addition to future predictions using our main CarFormer model with SAVi-14-light, we also provide future predictions and SAVi reconstructions from our model with SAVi-7-base.

In \subref{fig:ffcb} and \subref{fig:ffci}, the vehicle in the center of the scene is at an intersection. However, in the current timestep, the vehicle has yet to initiate the process of turning. Given only this information, the vehicle could end up taking either a right turn or a left turn. As a result, we see both models predicting a slight right turn while keeping the vehicle orientation the same. Due to the architecture of our model and the way future slot representations are predicted, we are not able to accurately capture multi-modality. As a result, the models predict the mean of two modes. This can also be seen with the vehicle on the left side in \subref{fig:ffcc} and \subref{fig:ffcj} making a right turn at the intersection. 

The inability of the SAVi-7-base model to capture vehicles in scenes with many vehicles is highlighted in \subref{fig:ffce} and \subref{fig:ffcg}. Due to the insufficient number of slots, the model outputs inaccurate predictions with blurry blobs. On the other hand, the SAVi-14-light model, with twice the number of slots, does not face the same problem as can be seen in  \subref{fig:ffcl} and \subref{fig:ffcn}. The model accurately captures individual vehicles, and subsequently, predicts accurate future states.

\section{Autoregressive Rollout with Block Attention}
\label{sec:rollout_gen}

Using block attention breaks the causality within these blocks, which is typically required for generating rollouts autoregressively and reliably. If we limit block attention to the training only, we cause a distribution shift during test time which is not ideal. As a result, we briefly outline here how we can autoregressively generate rollouts, which is necessary for certain RL approaches such as Trajectory Transformer \cite{Janner2021NeurIPS}. Note that to do this, we need to add a loss term to the loss function for next-word prediction, which is trivial as all inputs other than the scene representation are discretized.

In CarFormer, we use block attention exclusively in the scene representation part of the input. As this representation comes from another perception or feature extraction module, we can treat it as a unit. As a result, we assume we always begin with at least one set of features before we start generating rollouts. For the case of slots, at test time, we can start with the initial input trajectory as:

\begin{equation}
    \label{eq:traj_rollout}
    \tau_{inp} = \{g_t^x, g_t^y, l_t, v_t, \bz_t^1, \dots, \bz_t^K, \br_t^1, \br_t^2 \}
\end{equation}

We would like to rollout future trajectories given this initial context. In this specific trajectory, we use block attention within the scene representation $\{\bz_t^1, \dots, \bz_t^K, \br_t^1, \br_t^2\}$. 

To begin, we can autoregressively predict the actions for the current timestep $\{q_t^1, \dots, q_t^{2W}\}$ as well as the goal and initial part of the state of the next timestep $\{g_{t+1}^x, g_{t+1}^y, l_{t+1}, v_{t+1}\}$. This is a regular autoregressive next token prediction and does not require any modification. The only difference is that part of the input context contains block attention, which is similar to a variant introduced in \cite{Liu2018ICLR} with \textit{non-causal} attention within part of the prompt, typically the start of the prompt.

Next, to generate the slot features for time $t+1$, remember that we train our backbone with a forecasting objective for the scene-level representation at time $t+f$. As a result, we can do this if we set $f$ to 1 when training the CarFormer model. Given the forecasting head, we can get the next scene-level representation $\{\bz_{t+1}^1, \dots, \bz_{t+1}^K\}$. Moreover, we also need to predict the route representations for the next time step, which can be done in the same way we currently predict slots to get $\{\br_{t+1}^1, \br_{t+1}^2\}$. Although we currently limit the forecasting loss to only slot features, nothing is limiting us from also predicting these next route features by modifying the forecasting loss. Finally, as we know the entire representation for the next scene representation at time $t+1$, we can append it to the trajectory and continue autoregressively predicting the next time steps as required.

One downside of predicting the next time step in language modeling entirely in one shot as a block is that the individual predictions are not conditioned on each other, especially because the outputs in language are typically discrete tokens that are either greedily chosen or sampled according to the output probabilities from the language modeling head. However, since object-level features are continuous in our case, this is likely less of a problem as there is no sampling step and the features interact with each other up to the attention layer in the last transformer layer. For the case of VQ-VAE and other discretized scene representations, we refer the reader to Non-Autoregressive Transformers (NATs), which are proposed in \cite{Gu2018ICLR} for machine translation and other use cases.

\clearpage
\bibliographystyle{splncs04}
\bibliography{main}
\end{document}